\documentclass[twoside,11pt]{article}

\usepackage{jmlr2e}
\usepackage[cmex10]{amsmath}
\usepackage{amsmath,setspace,enumerate,amsbsy} %for ams math support% Definitions of handy macros can go
\DeclareMathAlphabet{\mathcal}{OMS}{cmsy}{m}{n}
\newcommand{\bs}[1]{\boldsymbol{#1}}
\firstpageno{1}

\begin{document}

%\title{Extended \textit{Rule-of-Thumb} for Bandwidth Selection in Univariate and Multivariate Kernel Density Estimation} %Methods}
\title{The Gram-Charlier A Series based Extended \textit{Rule-of-Thumb} for Bandwidth Selection in Univariate and Multivariate Kernel Density Estimations}
%Kernel Based  Univariate and Multivariate  Density and Density Derivative Estimations} %Methods}

\author{\name Dharmani Bhaveshkumar C. \hfill 				\email dharmanibc@gmail.com \\
       \addr Dhirubhai Ambani Institute of Technology (DAIICT), \\
       Gandhinagar, Gujarat, INDIA - 382001}

\editor{}

\maketitle

\begin{abstract}%   <- trailing '%' for backward compatibility of .sty file
The article derives a novel Gram-Charlier A (GCA) Series based Extended \textit{Rule-of-Thumb} (ExROT) for bandwidth selection in Kernel Density Estimation (KDE). 
There are existing various bandwidth selection rules achieving minimization of the Asymptotic Mean Integrated Square Error (AMISE) between the estimated probability density function (PDF) and the actual PDF.
%, is the most widely used performance criteria to derive various bandwidth selectio rules. 
%The rules differ in a way to estimate the required roughness of the second order derivative of the unknown PDF, where the roughness of a function is the integration of the squared function. 
The rules differ in a way to estimate the integration of the squared second order derivative of an unknown PDF  $(f(\cdot))$,  identified as the roughness $R(f''(\cdot))$. %  to optimize the AMISE. 
The simplest \textit{Rule-of-Thumb} (ROT) estimates the $R(f''(\cdot))$ %, to optimize the AMISE 
with an assumption that the density being estimated is Gaussian. 
Intuitively, better estimation of $R(f''(\cdot))$ and consequently better bandwidth selection rules 
can be derived, if the unknown PDF is approximated through an infinite series expansion based on a more generalized density assumption.  
%as an approximtion to unknown PDF to estimate $R(f''(\cdot))$ should bring better bandwidth selection rules. 
As a demonstration and  verification to this concept, the ExROT derived in the article uses an extended assumption  that the  density being estimated is near Gaussian. This helps use of 
%approximation of an unknown %probability density function 
%PDF through its higher order statistics based infinite series expansion to estimate the $R(f''(\cdot))$.
%% for optimizing the same AMISE criteria. 
%As an example, here, the Gram-Charlier A series 
the GCA expansion as an approximation to the unknown near Gaussian PDF. The ExROT for univariate KDE is extended to that for multivariate KDE. The required multivariate AMISE criteria is re-derived using  elementary calculus of several variables, instead of Tensor calculus. The derivation uses the Kronecker product and the vector differential operator to achieve the AMISE expression in vector notations.   
There is also derived ExROT for kernel based density derivative estimator. 
%The required multivariate Gram-Charlier expansion is also derived as an extension to a specific derivation of univariate Gram-Charlier expansion using multlinear Kronecker algebra. 
\end{abstract}


\section{Introduction}
\label{introduction}
%Kernel Density Estimation (KDE) %for continuous density function estimation 
%techniques  are  is widely used in statistics, machine learning and signal processing. The kernel selection and the bandwidth selection are the design parameters. 
Continuous probability density function (PDF) estimation using kernel methods is widely used in statistics, machine learning and signal processing \citep{Silverman86}. The optimal estimation depends upon the selected kernel function and its spread decided by the smoothing or bandwidth parameter. The selection of kernel has limited impact on optimal PDF estimation, although Epanechnikov kernel is the most optimal kernel  \citep{Epanechnikov69}. 
On the other hand, the optimal value of bandwidth parameter avoids either too rough or too smooth estimation of an  unknown PDF. %the consistancy and the rate of convergence of the density esimator  depends upon the bandwidth selection cite{Silverman78}.

\hspace{0.2 in} 	There exist variety of rules for bandwidth selection in KDE. The rules vary  based on the criteria to measure accuracy of  density estimation and to satisfy the used criteria.  The brief survey of data driven bandwidth selectors is provided by  \citet{wand1994kernel,bwKDEpark90,surveybwkde96,bwKDEpark92,Sheather92}. 
The Asymptotic Mean Integrated Square Error (AMISE) between the estimated PDF and the actual PDF is the most widely used performance criteria to derive the rules, though there are many others. 
The AMISE criteria requires estimating the roughness of the  squared second order PDF $(R(f''(\cdot)))$ as a prior step to estimate the kernel bandwidth parameter, where the roughness of a function is defined as integration of the squared function. The rules, based on the AMISE as a performance criteria, differ in a way to estimate the  $R(f''(\cdot))$. 
%%There are many methods to decide the bandwidth parameter. 
%The state of art in bandwidth estimation is to minimize the Asymptotic Mean Integrated Square Error (AMISE) between the estimated and true PDF. %The criteria requires to estimate second order derivative of the unknown density function.  
%The selectors mainly differ in a way to estimate second order derivative of the density function, required to minimize the AMISE criteria. 
The simplest \textit{Rule-of-Thumb (ROT)}, satisfying the AMISE, %for bandwidth selection in KDE
 by \citet{Silverman86} assumes Gaussian distribution for the unknown density. It is not the most optimal bandwidth selector but is used either as a very fast   reasonably good estimator or as a first estimator in multistage bandwidth selectors. More precise \textit{solve-the-equation plug-in} rules \citep{Sheather83,Sheather86}
 %satisfying the same  %based on AMISE criteria, 
 use estimation of integrated squared density derivative functionals to estimate $R(f''(\cdot))$. They demand high computations to solve a non-linear equation using iterative methods. They use ROT as a very first estimate. %, based on which the improvements are iteratively provded. 
 The fastest $\epsilon$-exact approximation algorithm based \textit{solve-the-equation plug-in} rule  \citep{Raykar05,Raykar06} requires $O(N+M)$ order of computations, where $N$ is number of samples and $M$ is the selected number of evaluation points. So, deriving a data dependent bandwidth parameter selection rule for KDE that balances accuracy and computation is the goal of this article. 
%There are also other \textit{plug-in} methods based on cross-validation and boot-nsstrap giving less optimal estimations using less computations. %ere are also available many  methods for bandwidth selection. 

\hspace{0.2 in} The article achieves this goal by deriving  an \textit{Extended Rule-of-Thumb} (ExROT). The assumption about Gaussianity of an unknown PDF in ROT is too restrictive. Expressing an unknown PDF, in terms of an infinite series of higher order statistics, based on a more generalized assumption should result into a better bandwidth selection rule.  %Better %PDF estimation could be derived if the 
%bandwidth selection rules can be derived if an unknowwn PDF is expressed as a higher order cumulants based infinite series expansion based on a more generalized assumption. 
As a verification and demonstration to this concept, the ExROT 
%is derived satisfying the same AMISE criteria with an 	
%extended assumption of near Gaussianity than the gassianity assumption of ROT. 
extends the Gaussian assumption in ROT to near Gaussian assumption. This facilitates use of cumulants based Gram-Charlier A (GCA) Series expansion as an  approximation for the unknown PDF to satisfy the same AMISE criteria. The empirical simulations prove that the ExROT for bandwidth selection is better than the ROT, with respect to an integrated mean square error (IMSE) or MISE performance criteria, for all types of nongaussian unimodal distributions including  the skewed, the kurtotic and with outlier distributions.  This is achieved 
%in case of unimodal, specifically the skewed and kurtotic, distributions 
with computation comparable to the ROT and too less compare to the \textit{$\epsilon$-exact solve-the-equation plug-in} rule. %just 6 multiplications and 6 additions after finding the cumulants. %Though little more comuptations than the ROT method, it requires almost 
%The PDF approximation through infinite series expansion is a well established area and there exist many such approximations. As the first results are encouraging, many such approximations be tried with Gaussian or other reference PDFs. Also, it seems possible to use ExROT to use as a first guess for the more precise \textit{solve-the-equation  plug-in} methods.

\hspace{0.2 in} The ExROT for bandwidth selection in univariate KDE is extended to the similar for multivariate KDE and   kernel based multivariate density derivative estimation. The ExROT for multivariate KDE requires  multivariate expression for AMISE,  multivariate Taylor Series expansion, multivariate Hermite polynomials and multivariate GCA Series expansion. 
The required multivariate AMISE is conventionally derived using gradient and Hessian of the PDF of a random vector \citep{wand1994kernel}. %, as  shown in Equation \eqref{oldndAMISE} in Section \ref{ndAMISE}. 
Conventionally, the other required multivariate expressions are also derived using Tensor calculus, as higher order derivatives of a multivariate functions are involved. Often, the corresponding final expressions requires coordinatewise representations. 
But recently, the higher order cumulants  \citep{ndHermite02,SankhyaCumVec06} and multivariate Hermite polynomials \citep{ndHermite02,dnHermite96} are derived  using only elementary calculus of several variables. This is achieved by replacing conventional multivariate differentiations by repeated applications of the Kronecker product to vector differential operator. The derived expressions are also more elementary as using vector notations and more comprehensive as apparently more nearer to their counterparts in univariate. The same approach has been used here to derive multivariate AMISE criteria in a vector notations. 
%using kronecker product.   
Overall, the multivariate ExROT is derived using the multivariate Taylor Series, the multivariate cumulants and the multivariate Hermite polynomials derived by \citet{ndHermite02,dnHermite96}, the multivariate GCA derived by \citet{ndGGCarxivDharmani} and  the multivariate AMISE obtained in Section \ref{ndAMISE} of this article. There is also derived bandwidth selection rule for kernel based density derivative estimation.   

%\hspace{0.2 in} The 1-dimensional bandwidth selector ExROT need be extended for multivariate, as it is to be used for least squares based kernel estimation of the independence contrasts. The extention will require Gram-Charlier series expansion for multivariate PDFs. One of the 
The next Section \ref{bwkde1dhistory} derives the univariate AMISE criteria and gives brief on the  existing rules for data driven kernel bandwidth selectors. The Section \ref{bwkde1dexrot} derives the ExROT. The performance analysis is done using two separate experiments in Section \ref{bwkde1dexp}. The preliminary background on multivariate KDE, Kronecker product, multivariate Taylor Series and others is briefed in Section \ref{ndprereq}. A compact derivation for multivariate GGC Series and GCA Series is provided in Section \ref{compactGGC}. 
Then, the Section \ref{ndAMISE} derives multivariate AMISE in a vector form using Kronecker Product. The multivariate AMISE and multivariate GCA Series are used to derive ExROT for multivariate KDE in Section \ref{ndExROT}. Similarly, the Section \ref{ExROT4Der} derives ExROT for density derivative estimation.   
Finally, the article is  concluded in Section \ref{bwkdeconclusion}. 
%cumulants based infinite series expansion approximation of the unknown density and derives \textit{extended or precise rule-of-thumb} equation for bandwidth selection. The method requires 6 multiplications and 6 additions after 
%finding the cumulants from the available data using $O(N)$ computations. [The performance is better than both the ROT method and equation based methods.] 
%[The method is extended for derivatives of the PDF estimation.]
%[In general, applications  in statistics need to find the density function estimation at a point. So, error in estimated value may be allowed. But, in case of machine learning, the maxima or  minima from a set of points is required. Then, using a same bandwidth parameter for all the different points may  affect badly the performance [cite kernel based correntropy results in ITL book]. The new method solves the problem by adapting the bandwidth to the given data point.] 
\section{Bandwidth Selection Criteria and Selectors}
\label{bwkde1dhistory}
 Given N realizations of an unknown PDF $f(x)$, the kernel density estimate ${\hat{f(x)}}$ is given by %The problem is to estimate $f^{^}(x)$. 
 \begin{equation}
 \label{eqkdech3}
 \hat{f(x)} = \frac{1}{Nh}\sum_{i=1}^{N}{K\left(\frac{x-x_i}{h}\right)} = \frac{1}{N}\sum_{i=1}^{N}{K_h\left(x-x_i\right)}
 \end{equation}
 where, $K(u)$ is the kernel function and h is the bandwidth parameter. Usually, $K(u)$ is a symmetric, positive definite and bounded function;  mostly a PDF;  satisfying the following properties:  %i.e. it satisfies the following properties:
 \begin{equation*}
 K(u) \geq 0, \mbox{  }\int_{-\infty}^{\infty}{K(u)du} = 1, \mbox{  }\int_{-\infty}^{\infty}{uK(u)Du} = 0, \mbox{  }\int_{-\infty}^{\infty}{u^2K(u)du} = \mu_2(K) < \infty 
 \end{equation*}
 The accuracy of a PDF estimation can be quantified by the available distance measures between PDFs; like; $L_1$ norm based mean integrated absolute measure, $L_2$ norm based mean integrated square error (MISE), Kullback-Libeler divergence and others. The optimal smoothing parameter (the bandwidth) $h$ is obtained by minimizing the selected distance measure. The most widely used criteria MISE or IMSE (Integrated Mean Square Error) based bandwidth selection rule, as in  \citep{Silverman86,wand1994kernel}, is briefed in the Appendix \ref{appamise1d}. It is given as under:
 \begin{eqnarray}
% \mbox{ISE}(f(x),{\hat{f(x)}}) &=& L_2(f(x),{\hat{f(x)}}) := \int_{-\infty}^{\infty}{({\hat{f(x)}}-f(x))^2 dx} \nonumber \\
 \mbox{MISE}({\hat{f(x)}} ) &=&  E\{ ISE( f(x),{\hat{f(x)}} )\} = E\left\{ \int_{-\infty}^{\infty}{ ( {\hat{f(x)}}-f(x) )^2 dx }\right\}  \nonumber \\
% &=&  \int_{-\infty}^{\infty}{ E\{ ({\hat{f(x)}}-f(x))^2 \} dx} = \int_{-\infty}^{\infty}{ \mbox{MSE}( f(x),{\hat{f(x)}} ) } = \mbox{IMSE}( f(x),{\hat{f(x)}} )  \nonumber \\
% &=& \int_{-\infty}^{\infty}{ ( E\{\hat{f(x)}\}-f(x) )^2 + E\{ ( {\hat{f(x)}} - E\{\hat{f(x)}\} )^2 \} dx}  \nonumber \\
\label{eqmise} 
 &=& \int_{-\infty}^{\infty}{\mbox{Bias}^2({\hat{f(x)}})dx} + \int_{-\infty}^{\infty}{\mbox{Var}({\hat{f(x)}})dx} \nonumber \\
 &=& \frac{h^4}{4}(\mu_2(K))^2R(f'') +  \frac{1}{Nh}R(K) + O(h^4) + O\left( \frac{h}{N} \right)
 \end{eqnarray}
where, $\mu_2(K) = \int{z^2K^2(z)dz}$,  
$R(f'') = \int{(f''(x))^2dx}$ and $R(K) = \int{K^2(z)dz}$. In general, $R(g) = \int{g^2(z)dz}$ is identified as the roughness of function $g(x)$. 
An asymptotic large sample approximation AMISE is obtained, assuming $\lim_{N\rightarrow\infty}{h} = 0$ and $\lim_{N\rightarrow\infty}{Nh} = \infty$ i.e. h reduces to 0 at a rate slower than $1/N$.
\begin{equation}
\label{eqmisef}
\mbox{AMISE}({\hat{f(x)}}) = \frac{h^4}{4}(\mu_2(K))^2R(f'') +  \frac{1}{Nh}R(K) 
\end{equation}
The Equation \eqref{eqmisef} interprets that a small $h$ increases estimation variance, whereas, a larger $h$ increases estimation bias.
% reduces estimation bias but increases estimation variance and a larger $h$ reduces estimation variance but  increases estimation bias.
An optimal $h$ minimizing the total $\mbox{AMISE}({\hat{f(x)}})$ is given by, 
\begin{eqnarray}
%\frac{d}{dh}AMISE({\hat{f(x)}}) &=& h^3(\mu_2(K))^2R(f'') -  \frac{1}{Nh^2}R(K) = 0  \nonumber \\
\label{eqhamise}
\Rightarrow h_{AMISE} &=& \left( \frac{R(k)}{\mu_2(K)^2R(f'')N}\right)^{\frac{1}{5}}
\end{eqnarray}
Thus, the optimal bandwidth parameter depends upon some of the kernel parameters, number of samples and the second derivative of the actual PDF being estimated. 
\subsection{Brief Survey on the bandwidth selectors}    %\hspace{0.2 in} 
As the bandwidth selection rules vary based on the choice of performance criteria for density estimation, they also vary based on the way the performance criteria is optimized. 
The various rules, satisfying AMISE, for bandwidth parameter selection differ in the way $R(f'')$ is estimated. The first group of  rules named \textit{scale measures} give rough estimate of the bandwidth parameter with less computation. It includes Silverman's  \textit{Rule-of-Thumb} that estimates $h$ assuming $f(x)$ being Gaussian \citep{Silverman86}. For a Gaussian PDF  $R(f'') = \frac{3\sigma^{-5}}{8\sqrt{\pi}}$ and for a Gaussian kernel $R(k) = \frac{1}{2\sqrt{\pi}}$. Accordingly, 
\begin{equation}
 h_{rot} = 1.0592\sigma N^{-1/5} 
\end{equation} 
where, $\sigma$ is the standard deviation of $f(x)$. There are many other rules based on the assumption of other parametric family. 
%Silverman gave another rule based on the interquantile range  (IQR) of the samples normalized by the IQR of normal distribution $N(0,1)$. 
There are also rules based on oversmoothed $h$, difference based $h$ and others briefed by \citep{janssen1995scale}. 
 
\hspace{0.2 in} An another group of rules is based on the more accurate at high computation \textit{plug-in} rules. They plug-in the kernel based estimate of the $R(f'')$. %The kernel based $f''(x)$ is obtained by differentiating equation \eqref{eqkdech3} twice with respect to x. But, the estimate also requires a bandwidth parameter. 
The \textit{direct plug-in} rules estimate derivative of the density functionals instead of estimating actual derivatives. Every $r^{th}$ order derivative functional estimation requires $(r+2)^{th}$ order estimate and pilot bandwidth to start with. Assuming, some parametric density for the $(r+2)^{th}$ order density the pilot bandwidth is selected and cumulatively the bandwidth parameter to estimate  $f(x)$ is obtained. The \textit{solve-the-equation plug-in} rules use the same approach but, instead of assuming bandwidth parameter, they optimize it by directly putting it into the AMISE. This requires solving a non-linear equation iteratively. They have better performances at high computation. 
Other than the rules for bandwidth selection, there are also cross-validation methods selecting the best from a user-defined list of bandwidth parameter based on some performance criteria. But, there is always a compromise between a length of the list for possible bandwidth parameters and the amount of computation.    

\hspace{0.2 in}	Over all, a bandwidth selection rule %that mingles advantages from both he groups 
that achieves precise bandwidth parameter at low computation is still open for research. 
%The more detail history o this class of methods  is found in \citep{wand1994kernel}. 
%\section{Cumulants Based Plug-In Method for Bandwidth Selection}
\section{Extended \textit{Rule-of-Thumb} (ExROT) Bandwidth Selector}
\label{bwkde1dexrot}
%As discussed in the previous section, conventional approaches to estimate $R(f'')$ are either assuming $f(x)$ from a parametric family or using it's kernel estimate. Another more intuitive approach to estimate $R(f'')$ would be by using cumulants based Gram-Charlier A-series expansion of $f(x)$. 
The Gaussianity assumption for an unknown PDF to estimate $R(f'')$ is too restrictive. Intuitively, better PDF estimations can be derived if $R(f'')$ is estimated by approximating PDFs through infinite series expansion. 
%Other than the conventional approaches, one more intuitive approach to estimate $R(f'')$ would be by approximating PDFs through infinite series expansion, 
As a verification and demonstration to this concept, ExROT is derived in this section using cumulants based Gram-Charlier A Series expansion of $f(x)$ based on near Gaussianity assumption for an unknown PDF $f(x)$. The series is given as: 
\begin{equation*}
f(x) = \frac{1}{\sqrt{2\pi}\sigma}\mbox{exp}\left[ \sum_{r=0	}^{\infty}{(k_r-\gamma_r)\frac{(-D)^r}{r!}}\right]\mbox{exp}\left( -\frac{1}{2}\left(\frac{x-\mu}{\sigma}\right)^2\right) 
\end{equation*}
where, $k_r$ is the $r^{th}$ cumulant of $f(x)$, $\gamma_r$ is the $r^{th}$ cumulant of Gaussian PDF (G(x)) and $D$ is the derivative operator with respect to x. 
%The expansion is derived using Taylor series of the characteristic function of $f(x)$ in terms of the characteristic function of a Gaussian, assuming near Gaussian PDF. The Taylor series expansion and it's second derivative are given as under:
%\begin{eqnarray*}
%f(x) &=& \sum_{n=0}^{\infty}{\frac{f^{(n)}(a)}{n!}(x-a)^n} \nonumber \\
%f''(x) &=& \sum_{n=0}^{\infty}{n(n-1)\frac{f^{(n)}(a)}{n!}(x-a)^{n-2}} \nonumber \\
% &=& \sum_{n=2}^{\infty}{\frac{f^{(n)}(a)}{(n-2)!}(x-a)^{n-2}} \nonumber \\
% &=& \sum_{n=0}^{\infty}{\frac{f^{(n+2)}(a)}{n!}(x-a)^{n}} \nonumber 
%\end{eqnarray*}
With $k_1=\gamma_1$, $k_2=\gamma_2$  and taking derivative twice with respect to x yields:
\begin{equation*}
f''(x) = \frac{1}{\sqrt{2\pi}\sigma}\mbox{exp}\left[ \sum_{r=0}^{\infty}{(k_{r}-\gamma_{r})\frac{(-D)^{(r+2)}}{r!}}\right]\mbox{exp}\left( -\frac{1}{2}\left(\frac{x-\mu}{\sigma}\right)^2\right) %\nonumber \\
\end{equation*}
Also, the nth derivative of Gaussian is given by
\begin{equation*}
 \frac{d^{(n)}G(x; \mu, \sigma)}{dx^n} = \frac{1}{\sigma^n}(-1)^nH_n(x)G(x; \mu,\sigma)
 \end{equation*}
where, $G(x)= \frac{1}{\sqrt{2\pi}\sigma}\exp\left[-\frac{1}{2}\left(\frac{x-\mu}{\sigma} \right)^2\right]$ is the Gaussian PDF and 
$H_n$ is the nth order Hermite polynomial. Accordingly, approximating upto fourth order cumulant 
\begin{eqnarray}
f''(x) &\approx &  \frac{1}{\sqrt{2\pi}\sigma}\mbox{exp}\left( -\frac{1}{2}\left(\frac{x-\mu}{\sigma}\right)^2\right)\left[ \frac{1}{\sigma^2}H_2(z) - \frac{k_3}{3!\sigma^5}H_5(z) + \frac{k_4}{4!\sigma^6}H_6(z) \right] \nonumber \\
\label{eqkf2}
R(f'') &=& \int_{\infty}^{\infty}{\frac{1}{2\pi\sigma}\mbox{exp}\left( -\left(z\right)^2\right)\left[ \frac{1}{\sigma^2}H_2(z) - \frac{k_3}{3!\sigma^5}H_5(z) + \frac{k_4}{4!\sigma^6}H_6(z) \right]^2dz} 
%\mbox{  }\left( \because z=\frac{x-\mu}{\sigma}\right)
\end{eqnarray}
The integration is obtained using the following rules: % of integration of Gaussian.
\begin{align*}
\int_{-\infty}^{\infty} \mbox{e}^{-a x^2}dx &= \sqrt{\frac{\pi}{a}} \\
\int_{-\infty}^{\infty} x^{2n} \mbox{e}^{-ax^2}dx &= 
\begin{cases}
\frac{(2n-1)!!}{(2a)^n}\sqrt{ \frac{\pi}{a} }  &\mbox{ for $n$ even} \\
0 & \mbox{ for $n$ odd, as the function becomes odd}
\end{cases}
\end{align*}
where, $(n-1)!! = (n-1)(n-3)(n-5)\ldots$.
Accordingly, the quantities in Equation \eqref{eqkf2} are obtained as under:
\begin{eqnarray}
T_1 &=& \int_{-\infty}^{\infty}{G^2(x)}\left[ \frac{1}{\sigma^2}H_2(x) \right]^2dx = \frac{1}{\sigma^5}\frac{3}{8\sqrt{\pi}} \nonumber \\
T_2 &=& \int_{-\infty}^{\infty}{G^2(x)}\left[ \frac{1}{\sigma^5}H_5(x) \right]^2dx = \frac{1}{\sigma^{11}}\frac{945}{64\sqrt{\pi}} \nonumber \\
T_3 &=& \int_{-\infty}^{\infty}{G^2(x)}\left[ \frac{1}{\sigma^6}H_6(x) \right]^2dx = \frac{1}{\sigma^{13}}\frac{10395}{128\sqrt{\pi}} \nonumber\\
T_4 &=& \int_{-\infty}^{\infty}{G^2(x)}\left[ \frac{2}{\sigma^7}H_2(x)H_5(x) \right]dx = 0 \nonumber \\
T_5 &=& \int_{-\infty}^{\infty}{G^2(x)}\left[ \frac{2}{\sigma^{11}}H_5(x)H_6(x) \right]dx = 0 \nonumber \\
T_6 &=& \int_{-\infty}^{\infty}{G^2(x)}\left[ \frac{2}{\sigma^8}H_2(x)H_6(x) \right]dx = \frac{1}{\sigma^9}\frac{105}{16\sqrt{\pi}} \nonumber 
\end{eqnarray}
This gives: 
\begin{eqnarray}
R(f'') &=& \frac{1}{\sigma^5}\frac{3}{8\sqrt{\pi}} \left[ 1 + \frac{1}{\sigma^{6}}\frac{315}{8}\left(\frac{k_3}{6}\right)^2 + \frac{1}{\sigma^{8}}\frac{3465}{16}\left(\frac{k_4}{24}\right)^2 + \frac{1}{\sigma^{4}}\frac{35}{2}\frac{k_4}{24}
\right] \nonumber \\
\label{eqrf2}
R(f'') &=& \frac{1}{\sigma^5}\frac{3}{8\sqrt{\pi}} \left[ 1 + 1.0938\frac{k_3^2}{\sigma^6} + 0.3764\frac{k_4^2}{\sigma^8}  + 0.7292 \frac{k_4}{\sigma^{4}}\right]
\end{eqnarray}
%For Gaussian kernel, $R\mu_2(K)/R(K) = 1/(2\sqrt{\pi})$. With this, 
Combining above Equation \eqref{eqrf2} with equation \eqref{eqhamise}, the Gram-Charlier A Series based an \textit{Extended Rule-of-Thumb} for bandwidth parameter $h_{GC}$ selection using near Gaussian PDF assumption and Gaussian kernel is given as under. As shown, the Silverman's \textit{Rule-of-Thumb} is one case of the extended rule.
% for some specific  and general cases:
\begin{eqnarray}
h_{cum} &=& 1.0592 \sigma (CN)^{-\frac{1}{5}} \\
\mbox{where,  }
C &=&
\begin{cases}
  1 & \text{both $k_3=K_4=0$ i.e. Gaussian PDF} \nonumber \\
 1 + 0.3764\frac{k_4^2}{\sigma^8} + 0.7292\frac{k_4}{\sigma^{4}} & \text{ if $k_3 = 0$ i.e. symmetric PDF}, \nonumber\\
 1 + 1.0938 k_3^2 + 0.3764 k_4^2 + 0.7292 k_4 	& \text{if $\sigma=1$}, \nonumber\\
% 1 + 2.9821k_4 	& \text{if both $k_3=0$ and $\sigma=1$}, \nonumber\\
 1 + 1.0938\frac{k_3^2}{\sigma^6} + 0.3764\frac{k_4^2}{\sigma^8} + 0.7292 \frac{k_4}{\sigma^{4}} & \text{otherwise} \nonumber
\end{cases}
%\begin{cases}
%1.0592\sigma \left(\left( 1 + 2.2559\frac{k_4}{\sigma^8} + 0.7292\frac{k_4}{\sigma^{4}}\right)N\right)^{-\frac{1}{5}} & \text{ if $k_3 = 0$ i.e. symmetric PDF}, \\
%1.0592 \left(\left( 1 + 6.5625 k_3 + 2.9821 k_4 \right)N\right)^{-\frac{1}{5}}	& \text{if $\sigma=1$}, \\
%1.0592 \left(\left( 1 + 2.9821k_4 \right)N\right)^{-\frac{1}{5}}	& \text{if both $k_3=0$ and $\sigma=1$}, \\
%1.0592\sigma \left(\left( 1 + 6.5625\frac{k_3}{\sigma^6} + 2.2559\frac{k_4}{\sigma^8} + 0.7292 \frac{k_4}{\sigma^{4}}\right)N\right)^{-\frac{1}{5}} & \text{otherwise}
%\end{cases}
\end{eqnarray}
%\section{Performance Analysis of Cumulant Based Bandwidth Parameter}
\section{Performance Analysis of ExROT Bandwidth Selector}
\label{bwkde1dexp}
\begin{figure}[!ht]
\includegraphics[scale=0.36]{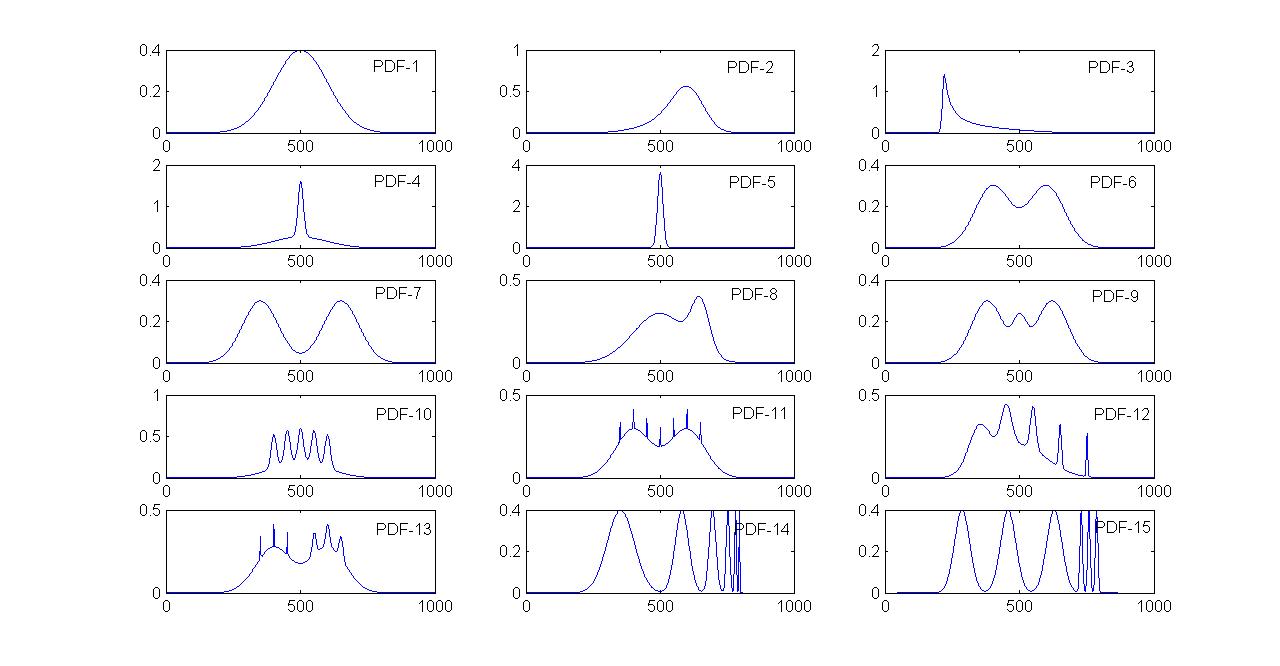} 
\centering
\caption{The Probability density functions (PDFs), generated through Normal mixtures, used to have performance comparison of various bandwidth selection rules for Kernel Density Estimation (KDE):  %against varying PDFs: 
(1) Gaussian (2) Skewed Unimodal (3) Strongly Skewed (4) Kurtotic Unimodal (5) Outlier (6) Bimodal (7) Separated Bimodal (8) Skewed Bimodal (9) Trimodal (10) Claw (11) Double Claw (12) Asymmetric Claw (13) asymmetric Double Claw (14) Smooth Comb (15) Discrete Comb } 
\label{bwkde1dpdf}
\end{figure}

There has been performed two separate experiments to test the performance of bandwidth selector. In both the experiments,  
the performance is tested on a set of 15 densities selected as a test-bed  for density estimators by  \citet{Marron92}. The  densities are shown in Figure \ref{bwkde1dpdf}. 
In both the experiments, the performance of ExROT is compared against Silverman's \textit{Rule-of-Thumb} and the $\epsilon$-exact approximation algorithm based \textit{solve-the-equation plug-in} rule \citep{Raykar05,Raykar06}.
\subsection{Experiment 1 (Performance against varying PDFs)} 
The first experiment is done to test the performance against varying PDFs. The experiment is done with 50000 samples and for 100 trials. The results  are shown in Table \ref{bwkde1dexp1}. Each table entry is an average of the 100 trials. 
The  selected three rules are compared for performances against three parameters - the value of bandwidth estimated, the corresponding IMSE %in density estimation 
between the estimated PDF and the theoretical PDF  %due to the selected bandwidth in each rule 
and the time taken in Seconds to estimate the bandwidth. 
%The IMSE is calculated taking the difference between the estimated PDF and the theoretical PDF from the normal mixture equation.
The theoretical PDF, required to calculate IMSE, is obtained from the normal mixture equations. That is why, all the 15  selected densities are generated using normal mixture equations.  

\hspace{0.2 in}		The \textit{$\epsilon$-exact solve-the-equation plug-in} rule is the best giving minimum IMSE error in all the cases accept the pure Gaussian density estimation. For Gaussian density,  ROT is slightly better than remaining both the rules.  The best IMSE performance of \textit{$\epsilon$-exact solve-the-equation plug-in} rule is at the cost of very high computation time. The mean time to estimate the bandwidth parameter for ROT is less than one millisecond. For ExROT it is about 10 to 20 milliseconds and the  same for \textit{$\epsilon$-exact solve-the-equation plug-in} rule is about 30 to 60 seconds. That means, the ExROT has  time complexity comparable to that of the ROT. So, the IMSE performance of these two needs a comparision. The boldface values for IMSE comparison in Table \ref{bwkde1dexp1}, show  a better between these two. It shows that in all non-Gaussian unimodal density estimation cases - skewed or kurtotic or with outlier - ExROT has outperformed ROT. The worst performance of ExROT in  multimodal density estimation is due to wrong estimation  of the  skewness and kurtosis. The ExROT  has also outperformed ROT  in some of the  -  claw and Asymmetric claw - multimodal density estimation cases. Thus, ExROT surely is a better option to ROT in unimodal density estimation. 
%It is also a promising option for multimodal density estimation,  as better approximation of kutosis and skewness or may be better  approximation of the multimodal densities through series expansion may improve further the performances. 
%% Thus,   a better multimodal PDF approximation through series expansion or   
\begin{table}[!ht]
\caption{Performance comparison of the bandwidth selection rules for Kernel Density Estimation (KDE) using 50000 samples. The results show the mean of the 100 trials. The boldface IMSE value show a better between the \textit{Rule-of-Thumb} (IMSErot) and \textit{extended Rule-of-Thumb} (IMSEexrot) rules. The time calculation is on a machine with features: Intel(R) Core(TM)2 Duo CPU, 2.93 GHz,  4.00 GB Internal RAM, 32 bit Windows 7 Professional, MATLAB R2010a} 
\centering
\footnotesize {
		\begin{tabular}{|c|ccc|ccc|ccc|} %|l
		\hline		
%\multicolumn{2}{|c|}{PDF} 	
\multicolumn{1}{|c|}{\bfseries PDF} &  \multicolumn{3}{c|}{Bandwidth (h)}  &  \multicolumn{3}{c|}{Integrated MSE (IMSE)*$10^5$}  & \multicolumn{3}{c|}{Estimation Time (in Sec)}\\ %\multicolumn{1}{|c|}{} & 
\cline{2-10}  %\hline
\bfseries Type%& \multicolumn{1}{c|}{\textbf{Type}} 
& \bfseries hrot & \bfseries hexrot & \bfseries heqfast & \bfseries rot & \bfseries exrot & \bfseries eqfast & \bfseries Trot & \bfseries Texrot & \bfseries Teqfast \\
\hline 
\hline
1 %& Gaussian 
&0.1217    & 0.1216    & 0.1213    & \bfseries 0.1424    & \bfseries 0.1424    & 0.1426    & 0.0005    & 0.0171   & 30.9182    \\
2 %& Skewed Unimodal 
 &  0.0993    & 0.0860    & 0.0818    & 0.1770    & \bfseries 0.1702    &  0.1702    & 0.0005    & 0.0157   & 35.9430 \\
3  %& Strongly Skewed
  & 0.1263    & 0.0964    & 0.0200    & 3.7034    &\bfseries  2.8516    & 0.4061    & 0.0006    & 0.0155   & 52.1223 \\
4 %& Kurtotic Unimodal 
&     0.0996    & 0.0693    & 0.0204    & 3.0476    & \bfseries 1.8294    & 0.3834    & 0.0008    & 0.0154   & 48.2712  \\
5 %& Outlier  
 &   0.0402    & 0.0044    & 0.0129    & 1.9249    & \bfseries 0.7042    & 0.4425    & 0.0006    & 0.0148   & 41.9652       \\
6 %&  Bimodal    
 &     0.1464    & 0.1632    & 0.0967    & \bfseries 0.1908    & 0.2220    & 0.1504    & 0.0006    & 0.0149   & 41.0530 \\
7 %& Separated Bimodal    
  &    0.1999    & 0.2065    & 0.0925    & \bfseries 0.3681    & 0.3897    & 0.1640    & 0.0006    & 0.0147   & 40.6588  \\
8 %& Skewed Bimodal  
 &   0.1333    & 0.1596    & 0.0736    & \bfseries 0.3122    & 0.4117    & 0.1920    & 0.0006    & 0.0153   & 41.7342  \\
9 %& Trimodal 
 &   0.1552    & 0.1700    & 0.0785    & \bfseries 0.3438    & 0.3954    & 0.1781    & 0.0005    & 0.0145   & 41.1552    \\
10 %& Claw  
&    0.1058    & 0.1042    & 0.0242    & 2.3709    &\bfseries  2.3269    & 0.3392    & 0.0005    & 0.0149   & 48.2730   \\
11 %& Double Claw 
&     0.1459    & 0.1629    & 0.0895    & \bfseries 0.7810    & 0.7961    & 0.7357    & 0.0005    & 0.0146   & 41.7426  \\
12  %& Asymmetric Claw 
& 0.1356    & 0.1350    & 0.0323    & 1.6634    &\bfseries  1.6586    & 0.5287    & 0.0005    & 0.0156   & 54.3546 \\
13 %& Asym DoubleClaw 
&     0.1450    & 0.1652    & 0.0461    & \bfseries 1.1161    & 1.1926    & 0.5791    & 0.0005    & 0.0147   & 47.9025   \\
14  %& Smooth Comb 
 &      0.2001    & 0.2058    & 0.0280    & \bfseries 3.2280    & 3.2763    & 0.8810    & 0.0005    & 0.0140   & 51.0985       \\
15  %& Discrete Comb 
&  0.2059    & 0.2111    & 0.0230    & \bfseries 3.2382    & 3.2824    & 0.4599    & 0.0006    & 0.0145   & 50.8321    \\
\hline
\end{tabular}
\label{bwkde1dexp1} 
}
\end{table}%\clearpage

%\section{Performance Analysis of Cumulant Based Bandwidth Parameter}
%The performance is tested on a set of 15 densities given as a test-bed  for density estimators in article \citep{Marron92}. The performance is compared against Silverman's \textit{Rule-of-Thumb} method and a \textit{solve the equation plug-in} method with fast computations discussed in \citep{Raykar06}.
\subsection{Experiment 2 (Performance against varying number of samples)}
\begin{figure}[!ht]
\includegraphics[scale=0.4]{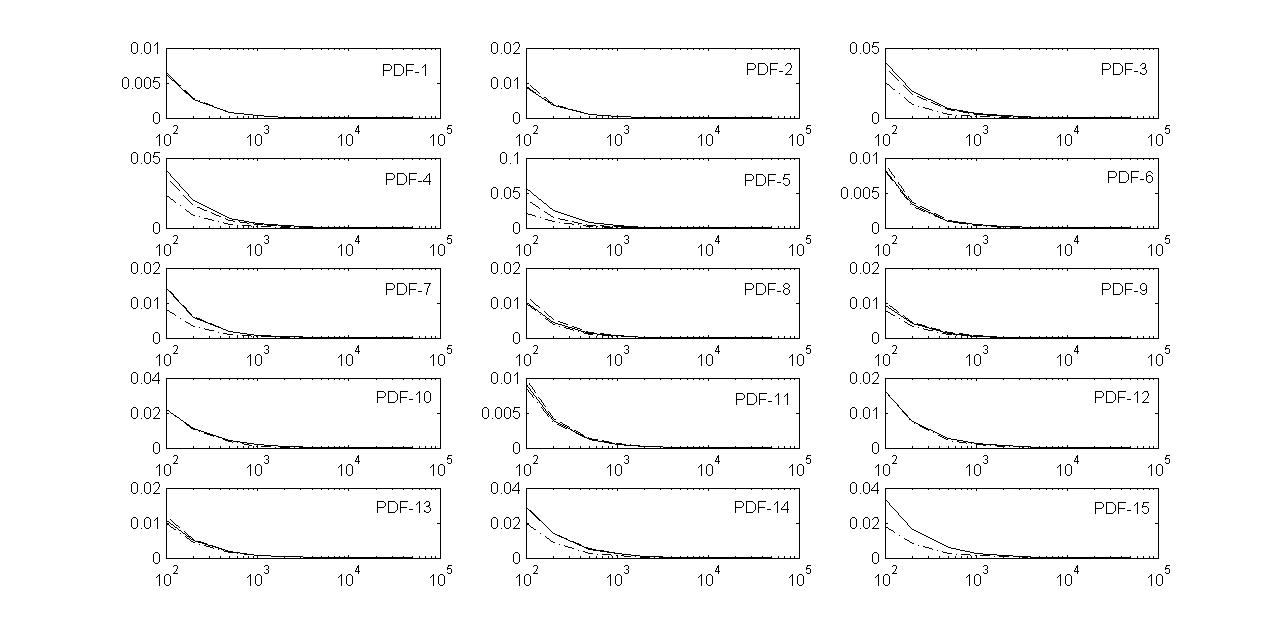} 
\centering
\caption{The Integrated Mean Square Error (IMSE) comparision of the bandwidth selection rules for Kernel Density Estimation (KDE) of  various PDFs against varying number of samples. The solid lines (-) indicate  \textit{Rule-of-Thumb}; dashed lines (- -) indicate \textit{extended Rule-of-Thumb}; the dash-dot lines (-.) indicate the \textit{$\epsilon$-exact solve-the-equation plug-in} rule }
\label{bwkde1dimserr}
\end{figure}

The second experiment is done to have the performance comparision of the same selected three bandwidth estimators against varying number of samples. The results of estimated bandwidth parameter, the IMSE and the estimation time (in Seconds) versus number of samples for varying PDFs are tabulated in Table \ref{bwkde1dexp21}. For better interpretations, the IMSE versus log of the number of samples are plotted; for all 15 PDFs and number of samples varying from 100 to the 50000; as shown in Figure \ref{bwkde1dimserr}.

\hspace{0.2 in}	The IMSE performances against varying number of samples (Nsamples) for varying PDFs are similar to that discussed in Experiment 1. The ExROT performance is better for unimodal skewed, kurtotic or with outlier densities. Even it is  better in some cases of multimodal skewed densities. Also, for small number of samples the ExROT performance is 
better compare to ROT. The convergence performance of ExROT is  matching other two rules. The IMSE decreases almost inversely with increase in number of samples.
\begin{table}[!htbp]
\caption{Performance comparison of the bandwidth selection rules for Kernel Density Estimation (KDE) against varying number of samples. The results show the mean of the 50 trials. The time calculation is on a machine with features: Intel(R) Core(TM)2 Duo CPU, 2.93 GHz,  4.00 GB Internal RAM, 32 bit Windows 7 Professional, MATLAB R2010a} 
\centering
\footnotesize {
		\begin{tabular}{|c|c|ccc|ccc|ccc|}
		\hline		
\multicolumn{1}{|c|}{\textbf PDF}& \bfseries Nsamples &  \multicolumn{3}{c|}{Bandwidth Parameter (h)} &  \multicolumn{3}{c|}{Integrated MSE (IMSE)*$10^3$} & \multicolumn{3}{c|}{Estimation Time (in Sec)}\\
\cline{3-11}
\bfseries Type & \bfseries *$10^-3$ & \bfseries hrot & \bfseries hexrot & \bfseries heqfast & \bfseries rot & \bfseries exrot & \bfseries eqfast & \bfseries Trot & \bfseries Texrot & \bfseries Teqfast \\
\hline 
\hline
 & 0.0100    & 0.4189    & 0.4379    & 0.4059    & 6.2430    & 6.5102    & 6.5579    & 0.0011    & 0.0004    & 0.0968\\
    & 0.0200    & 0.3613    & 0.3797    & 0.3573    & 2.6582    & 2.6653    & 2.7344    & 0.0001    & 0.0002    & 0.1210\\
    & 0.0500    & 0.3038    & 0.3099    & 0.3008    & 0.7831    & 0.7855    & 0.8064    & 0.0001    & 0.0004    & 0.3287\\
    & 0.1000    & 0.2659    & 0.2681    & 0.2602    & 0.3148    & 0.3176    & 0.3239    & 0.0001    & 0.0006    & 0.6734\\
   1 & 0.2000    & 0.2324    & 0.2337    & 0.2287    & 0.1253    & 0.1262    & 0.1274    & 0.0001    & 0.0010    & 1.3470\\
    & 0.5000    & 0.1929    & 0.1931    & 0.1911    & 0.0346    & 0.0346    & 0.0349    & 0.0001    & 0.0024    & 3.2634\\
    & 1.0000    & 0.1680    & 0.1678    & 0.1672    & 0.0127    & 0.0128    & 0.0128    & 0.0002    & 0.0038    & 6.3346\\
    & 2.0000    & 0.1464    & 0.1466    & 0.1455    & 0.0050    & 0.0050    & 0.0050    & 0.0002    & 0.0073   & 12.8339\\
%    & 5.0000    & 0.1217    & 0.1216    & 0.1213    & 0.0014    & 0.0014    & 0.0014    & 0.0005    & 0.0171   & 30.9182\\
\hline    
    & 0.0100    & 0.3454    & 0.3723    & 0.2944    & 8.9172   & 10.3399    & 9.2006    & 0.0001    & 0.0002    & 0.0720\\
    & 0.0200    & 0.2993    & 0.3130    & 0.2533    & 3.4302    & 3.8549    & 3.4604    & 0.0001    & 0.0002    & 0.1420\\
    & 0.0500    & 0.2465    & 0.2339    & 0.2026    & 0.9394    & 0.9924    & 0.9915    & 0.0001    & 0.0004    & 0.3852\\
    & 0.1000    & 0.2165    & 0.1870    & 0.1819    & 0.3954    & 0.3974    & 0.3945    & 0.0001    & 0.0007    & 0.7551\\
2    & 0.2000    & 0.1893    & 0.1663    & 0.1565    & 0.1399    & 0.1374    & 0.1381    & 0.0001    & 0.0010    & 1.5336\\
    & 0.5000    & 0.1571    & 0.1367    & 0.1296    & 0.0425    & 0.0422    & 0.0422    & 0.0001    & 0.0023    & 3.7511\\
    & 1.0000    & 0.1370    & 0.1198    & 0.1134    & 0.0168    & 0.0162    & 0.0161    & 0.0002    & 0.0036    & 7.2034\\
    & 2.0000    & 0.1194    & 0.1038    & 0.0986    & 0.0066    & 0.0064    & 0.0064    & 0.0003    & 0.0065   & 14.2614\\
 %   & 5.0000    & 0.0993    & 0.0860    & 0.0818    & 0.0018    & 0.0017    & 0.0017    & 0.0005    & 0.0157   & 35.9430\\
\hline 
    & 0.0100    & 0.4429    & 0.3498    & 0.1650   & 40.3373   & 36.7614   & 25.9410    & 0.0001    & 0.0002    & 0.0864\\
    & 0.0200    & 0.3846    & 0.2996    & 0.1188   & 18.8112   & 16.7646    & 9.6509    & 0.0001    & 0.0002    & 0.1738\\
    & 0.0500    & 0.3183    & 0.2459    & 0.0844    & 6.8916    & 6.0170    & 2.8246    & 0.0001    & 0.0004    & 0.4436\\
    & 0.1000    & 0.2775    & 0.2125    & 0.0653    & 3.2038    & 2.7464    & 1.0352    & 0.0001    & 0.0006    & 0.9069\\
 3   & 0.2000    & 0.2410    & 0.1845    & 0.0510    & 1.4770    & 1.2452    & 0.3778    & 0.0001    & 0.0010    & 1.9033\\
    & 0.5000    & 0.2003    & 0.1532    & 0.0384    & 0.5282    & 0.4362    & 0.1080    & 0.0001    & 0.0021    & 4.9425\\
    & 1.0000    & 0.1744    & 0.1332    & 0.0310    & 0.2397    & 0.1939    & 0.0377    & 0.0002    & 0.0032   & 10.3360\\
    & 2.0000    & 0.1518    & 0.1158    & 0.0257    & 0.1080    & 0.0856    & 0.0151    & 0.0003    & 0.0058   & 21.1275\\
 %   & 5.0000    & 0.1263    & 0.0964    & 0.0200    & 0.0370    & 0.0285    & 0.0041    & 0.0006    & 0.0155   & 52.1223\\
\hline
    & 0.0100    & 0.3500    & 0.2631    & 0.1247   & 41.5059   & 36.6378   & 23.6608    & 0.0001    & 0.0002    & 0.1076\\
    & 0.0200    & 0.2999    & 0.2169    & 0.0934   & 19.3056   & 16.1740    & 8.6166    & 0.0001    & 0.0002    & 0.2175 \\
    & 0.0500    & 0.2517    & 0.1805    & 0.0686    & 7.0407    & 5.6630    & 2.3757    & 0.0001    & 0.0003    & 0.5464\\
    & 0.1000    & 0.2167    & 0.1509    & 0.0557    & 3.2218    & 2.4513    & 0.9320    & 0.0001    & 0.0006    & 1.0900\\
 4   & 0.2000    & 0.1899    & 0.1333    & 0.0449    & 1.4626    & 1.0710    & 0.3232    & 0.0001    & 0.0010    & 2.1309\\
    & 0.5000    & 0.1582    & 0.1102    & 0.0350    & 0.5031    & 0.3439    & 0.0875    & 0.0001    & 0.0022    & 4.8694\\
    & 1.0000    & 0.1375    & 0.0957    & 0.0296    & 0.2205    & 0.1448    & 0.0344    & 0.0002    & 0.0037    & 9.6654\\
    & 2.0000    & 0.1197    & 0.0833    & 0.0252    & 0.0951    & 0.0600    & 0.0136    & 0.0002    & 0.0058   & 19.3117\\
  %  & 5.0000    & 0.0996    & 0.0693    & 0.0204    & 0.0305    & 0.0183    & 0.0038    & 0.0008    & 0.0154   & 48.2712\\
\hline
    & 0.0100    & 0.1329    & 0.0158    & 0.0511   & 56.5031   & 41.3602   & 21.4029    & 0.0003    & 0.0003    & 0.1138\\
    & 0.0200    & 0.1215    & 0.0145    & 0.0429   & 25.6093   & 14.6056    & 8.5744    & 0.0001    & 0.0002    & 0.2072\\
    & 0.0500    & 0.0992    & 0.0110    & 0.0352    & 8.1567    & 4.2720    & 2.5379    & 0.0001    & 0.0003    & 0.4943\\
    & 0.1000    & 0.0881    & 0.0100    & 0.0297    & 3.4558    & 1.6044    & 0.9628    & 0.0001    & 0.0006    & 0.9657\\
  5  & 0.2000    & 0.0761    & 0.0084    & 0.0257    & 1.4268    & 0.6472    & 0.4009    & 0.0001    & 0.0010    & 1.7802\\
    & 0.5000    & 0.0640    & 0.0072    & 0.0209    & 0.4287    & 0.1696    & 0.1052    & 0.0001    & 0.0017    & 4.2263\\
    & 1.0000    & 0.0556    & 0.0062    & 0.0181    & 0.1705    & 0.0652    & 0.0397    & 0.0002    & 0.0028    & 8.4271\\
    & 2.0000    & 0.0483    & 0.0053    & 0.0156    & 0.0675    & 0.0244    & 0.0156    & 0.0002    & 0.0056   & 16.8823\\
  %  & 5.0000    & 0.0402    & 0.0044    & 0.0129    & 0.0192    & 0.0070    & 0.0044    & 0.0006    & 0.0148   & 41.9652\\

\hline
\end{tabular}
\label{bwkde1dexp21} 
}
\end{table}%\clearpage

\begin{table}[!htbp]
\caption{Performance comparison of the smoothing (bandwidth) parameter selection rules for Kernel Density Estimation (KDE) against varying number of samples. The results show the mean of the 50 trials. The time calculation is on a machine with features: Intel(R) Core(TM)2 Duo CPU, 2.93 GHz,  4.00 GB Internal RAM, 32 bit Windows 7 Professional, MATLAB R2010a} 
\centering
\footnotesize {
		\begin{tabular}{|c|c||c|c|c||c|c|c||c|c|c|}
		\hline		
\multicolumn{1}{|c|}{\textbf PDF}& \bfseries Nsamples &  \multicolumn{3}{c||}{Bandwidth Parameter (h)} &  \multicolumn{3}{|c||}{Integrated MSE (IMSE)*$10^3$} & \multicolumn{3}{|c|}{h Calculation Time (T)}\\
\cline{3-11}
\bfseries Type& \bfseries *$10^-3$ & \bfseries hrot & \bfseries hexrot & \bfseries heqfast & \bfseries rot & \bfseries exrot & \bfseries eqfast & \bfseries Trot & \bfseries Texrot & \bfseries Teqfast \\
\hline 
\hline
 & 0.0100    & 0.5128    & 0.5781    & 0.3875    & 8.3749    & 9.1194    & 8.1524    & 0.0001    & 0.0002    & 0.0670\\
    & 0.0200    & 0.4421    & 0.4948    & 0.3186    & 3.2862    & 3.6643    & 3.0184    & 0.0001    & 0.0002    & 0.1404\\
    & 0.0500    & 0.3652    & 0.4096    & 0.2540    & 0.9741    & 1.1004    & 0.8728    & 0.0001    & 0.0004    & 0.3652\\
    & 0.1000    & 0.3190    & 0.3558    & 0.2254    & 0.4326    & 0.4840    & 0.3629    & 0.0001    & 0.0006    & 0.7373\\
 6  & 0.2000    & 0.2780    & 0.3105    & 0.1927    & 0.1644    & 0.1873    & 0.1335    & 0.0001    & 0.0010    & 1.5391\\\
    & 0.5000    & 0.2315    & 0.2583    & 0.1558    & 0.0483    & 0.0550    & 0.0402    & 0.0001    & 0.0019    & 3.9894\\
    & 1.0000    & 0.2020    & 0.2251    & 0.1356    & 0.0186    & 0.0214    & 0.0145    & 0.0002    & 0.0028    & 8.0967\\\
    & 2.0000    & 0.1759    & 0.1960    & 0.1174    & 0.0072    & 0.0083    & 0.0056    & 0.0003    & 0.0054   & 16.3610\\
  %  & 5.0000    & 0.1464    & 0.1632    & 0.0967    & 0.0019    & 0.0022    & 0.0015    & 0.0006    & 0.0149   & 41.0530\\
 \hline   
    & 0.0100    & 0.6923    & 0.7155    & 0.3566   & 14.2086   & 14.7060    & 8.3466    & 0.0001    & 0.0002    & 0.0742\\
    & 0.0200    & 0.5997    & 0.6202    & 0.3015    & 5.8889    & 6.1350    & 3.1726    & 0.0001    & 0.0002    & 0.1494\\
    & 0.0500    & 0.5005    & 0.5175    & 0.2410    & 1.7750    & 1.8618    & 0.8813    & 0.0001    & 0.0003    & 0.3908\\
    & 0.1000    & 0.4367    & 0.4512    & 0.2086    & 0.7399    & 0.7760    & 0.3704    & 0.0001    & 0.0006    & 0.7544\\
  7  & 0.2000    & 0.3807    & 0.3933    & 0.1800    & 0.2883    & 0.3037    & 0.1360    & 0.0001    & 0.0010    & 1.5005\\
    & 0.5000    & 0.3167    & 0.3272    & 0.1489    & 0.0854    & 0.0901    & 0.0397    & 0.0001    & 0.0018    & 3.8518\\
    & 1.0000    & 0.2759    & 0.2850    & 0.1285    & 0.0321    & 0.0340    & 0.0141    & 0.0002    & 0.0031    & 7.9127\\
    & 2.0000    & 0.2402    & 0.2481    & 0.1117    & 0.0130    & 0.0138    & 0.0058    & 0.0002    & 0.0054   & 16.0277\\
   % & 5.0000    & 0.1999    & 0.2065    & 0.0925    & 0.0037    & 0.0039    & 0.0016    & 0.0006    & 0.0147   & 40.6588\\
 \hline  
    & 0.0100    & 0.4647    & 0.5979    & 0.3305   & 10.3872   & 11.9845    & 9.8866    & 0.0001    & 0.0002    & 0.0731\\\
    & 0.0200    & 0.4010    & 0.5166    & 0.2932    & 4.3730    & 5.1674    & 3.8251    & 0.0001    & 0.0002    &0.1431\\
    & 0.0500    & 0.3390    & 0.4076    & 0.2206    & 1.4163    & 1.6669    & 1.1203    & 0.0001    & 0.0004    & 0.4028\\
    & 0.1000    & 0.2907    & 0.3550    & 0.1834    & 0.5798    & 0.7038    & 0.4435    & 0.0001    & 0.0007    & 0.8379\\
   8 & 0.2000    & 0.2538    & 0.3065    & 0.1534    & 0.2290    & 0.2850    & 0.1614    & 0.0001    & 0.0010    & 1.7051\\
    & 0.5000    & 0.2114    & 0.2527    & 0.1225    & 0.0674    & 0.0856    & 0.0441    & 0.0001    & 0.0019    & 4.1356\\
    & 1.0000    & 0.1841    & 0.2201    & 0.1045    & 0.0268    & 0.0345    & 0.0173    & 0.0002    & 0.0030    & 8.2747\\
    & 2.0000    & 0.1602    & 0.1917    & 0.0896    & 0.0106    & 0.0139    & 0.0068    & 0.0002    & 0.0054   & 16.6340\\
    %& 5.0000    & 0.1333    & 0.1596    & 0.0736    & 0.0031    & 0.0041    & 0.0019    & 0.0006    & 0.0153   & 41.7342\\\
 \hline
    & 0.0100    & 0.5406    & 0.5921    & 0.3817    & 9.4215   & 10.1084    & 7.9126    & 0.0003    & 0.0002    & 0.0782\\
    & 0.0200    & 0.4662    & 0.5133    & 0.3161    & 4.0153    & 4.3435    & 3.4017    & 0.0001    & 0.0002    & 0.1410\\
    & 0.0500    & 0.3881    & 0.4268    & 0.2522    & 1.3227    & 1.4426    & 1.0262    & 0.0001    & 0.0003    & 0.3687\\
    & 0.1000    & 0.3397    & 0.3718    & 0.2046    & 0.5546    & 0.6094    & 0.3941    & 0.0001    & 0.0007    & 0.7998\\
9    & 0.2000    & 0.2952    & 0.3235    & 0.1741    & 0.2397    & 0.2638    & 0.1620    & 0.0001    & 0.0010    & 1.6285\\
    & 0.5000    & 0.2461    & 0.2696    & 0.1354    & 0.0713    & 0.0800    & 0.0426    & 0.0001    & 0.0018    & 4.1529\\
    & 1.0000    & 0.2143    & 0.2345    & 0.1148    & 0.0293    & 0.0330    & 0.0168    & 0.0002    & 0.0030    & 8.1136\\
    & 2.0000    & 0.1865    & 0.2042    & 0.0979    & 0.0119    & 0.0135    & 0.0066    & 0.0002    & 0.0055   & 16.2836\\
    %5.0000    0.1552    0.1700    0.0785    0.0034    0.0040    0.0018    0.0005    0.0145   41.1552\\
\hline
    & 0.0100    & 0.3648    & 0.3635    & 0.3384   & 22.3775   & 22.4492   & 22.3534    & 0.0001    & 0.0002    & 0.0631\\
    & 0.0200    & 0.3172    & 0.3375    & 0.2825   & 10.9313   & 10.9990   & 10.8045    & 0.0001    & 0.0002    & 0.1406\\
    & 0.0500    & 0.2635    & 0.2995    & 0.1727    & 4.2370    & 4.2810    & 3.5900    & 0.0001    & 0.0004    & 0.6232\\
    & 0.1000    & 0.2304    & 0.2471    & 0.0805    & 2.0575    & 2.0706    & 0.9325    & 0.0001    & 0.0006    & 1.0865\\
 10   & 0.2000    & 0.2011    & 0.2032    & 0.0563    & 0.9881    & 0.9879    & 0.3128    & 0.0001    & 0.0010    & 2.0001\\
    & 0.5000    & 0.1675    & 0.1668    & 0.0427    & 0.3624    & 0.3609    & 0.0834    & 0.0001    & 0.0018    & 5.3371\\\
    & 1.0000    & 0.1459    & 0.1454    & 0.0359    & 0.1648    & 0.1642    & 0.0329    & 0.0002    & 0.0028   & 10.9701\\
    & 2.0000    & 0.1271    & 0.1254    & 0.0303    & 0.0728    & 0.0719    & 0.0124    & 0.0002    & 0.0055   & 21.3417\\
   % & 5.0000    & 0.1058    & 0.1042    & 0.0242    & 0.0237    & 0.0233    & 0.0034    & 0.0005    & 0.0149   & 48.2730\\
\hline
\end{tabular}
\label{bwkde1dexp2} }
\end{table}%\clearpage

\begin{table}[!htbp]
\caption{Performance comparison of the smoothing (bandwidth) parameter selection rules for Kernel Density Estimation (KDE) against varying number of samples. The results show the mean of the 50 trials. The time calculation is on a machine with features: Intel(R) Core(TM)2 Duo CPU, 2.93 GHz,  4.00 GB Internal RAM, 32 bit Windows 7 Professional, MATLAB R2010a} 
\centering
\footnotesize {
		\begin{tabular}{|c|c||c|c|c||c|c|c||c|c|c|}
		\hline		
\multicolumn{1}{|c|}{\textbf PDF}& \bfseries Nsamples &  \multicolumn{3}{c||}{Bandwidth Parameter (h)} &  \multicolumn{3}{|c||}{Integrated MSE (IMSE)*$10^3$} & \multicolumn{3}{|c|}{h Calculation Time (T)}\\
\cline{3-11}
\bfseries Type& \bfseries *$10^-3$ & \bfseries hrot & \bfseries hexrot & \bfseries heqfast & \bfseries rot & \bfseries exrot & \bfseries eqfast & \bfseries Trot & \bfseries Texrot & \bfseries Teqfast \\
\hline 
\hline
   
   & 0.0100    & 0.5102    & 0.5779    & 0.3879    & 9.1443    & 9.9017    & 8.6536    & 0.0001    & 0.0002    & 0.0664\\
    & 0.0200    & 0.4393    & 0.4940    & 0.3334    & 3.9503    & 4.2723    & 3.6995    & 0.0001    & 0.0002    & 0.1355\\
    & 0.0500    & 0.3681    & 0.4133    & 0.2540    & 1.2492    & 1.3551    & 1.1489    & 0.0001    & 0.0003    & 0.3606\\
    & 0.1000    & 0.3193    & 0.3569    & 0.2210    & 0.5709    & 0.6103    & 0.5283    & 0.0001    & 0.0006    & 0.7447\\
    11& 0.2000    & 0.2778    & 0.3103    & 0.1894    & 0.2513    & 0.2663    & 0.2331    & 0.0001    & 0.0010    & 1.5428\\
    & 0.5000    & 0.2311    & 0.2585    & 0.1560    & 0.0902    & 0.0941    & 0.0848    & 0.0001    & 0.0018    & 3.9834\\
    & 1.0000    & 0.2013    & 0.2250    & 0.1332    & 0.0422    & 0.0435    & 0.0399    & 0.0002    & 0.0028    & 8.1323\\
    & 2.0000    & 0.1754    & 0.1958    & 0.1140    & 0.0204    & 0.0210    & 0.0193    & 0.0003    & 0.0055   & 16.4165\\
    %& 5.0000    & 0.1459    & 0.1629    0.0895    0.0078    0.0080    0.0074    0.0005    0.0146   41.7426\\
\hline
    & 0.0100    & 0.4851    & 0.4874    & 0.4530   & 16.3492   & 16.3720   & 16.2242    & 0.0001    & 0.0002    & 0.0590\\
    & 0.0200    & 0.4049    & 0.4045    & 0.3320    & 7.6083    & 7.6088    & 7.1462    & 0.0001    & 0.0002    & 0.1398\\
    & 0.0500    & 0.3403    & 0.3400    & 0.1940    & 2.8506    & 2.8496    & 2.2314    & 0.0001    & 0.0003    & 0.4242\\
    & 0.1000    & 0.2963    & 0.2950    & 0.1373    & 1.3352    & 1.3324    & 0.9125    & 0.0001    & 0.0006    & 0.8973\\
 12   & 0.2000    & 0.2583    & 0.2578    & 0.1070    & 0.6246    & 0.6240    & 0.3869    & 0.0001    & 0.0010    & 1.8709\\
    & 0.5000    & 0.2151    & 0.2138    & 0.0748    & 0.2231    & 0.2223    & 0.1165    & 0.0001    & 0.0018    & 5.2212\\
    & 1.0000    & 0.1871    & 0.1864    & 0.0583    & 0.1022    & 0.1020    & 0.0480    & 0.0002    & 0.0028   & 10.5090\\
    & 2.0000    & 0.1628    & 0.1620    & 0.0449    & 0.0469    & 0.0468    & 0.0189    & 0.0002    & 0.0053   & 21.5027\\
    %5.0000    0.1356    0.1350    0.0323    0.0166    0.0166    0.0053    0.0005    0.0156   54.3546\\
\hline    
    & 0.0100    & 0.5067    & 0.5846    & 0.3639   & 10.9392   & 11.8958   & 10.0277    & 0.0009    & 0.0002    & 0.0855\\
    & 0.0200    & 0.4336    & 0.5002    & 0.3073    & 4.8686    & 5.2675    & 4.5417    & 0.0001    & 0.0002    & 0.1442\\
    & 0.0500    & 0.3617    & 0.4161    & 0.2481    & 1.7277    & 1.8489    & 1.5896    & 0.0001    & 0.0004    & 0.3750\\
    & 0.1000    & 0.3177    & 0.3619    & 0.2087    & 0.7961    & 0.8408    & 0.7279    & 0.0001    & 0.0006    & 0.7774\\
   13 & 0.2000    & 0.2758    & 0.3146    & 0.1767    & 0.3753    & 0.3933    & 0.3341    & 0.0001    & 0.0010    & 1.6145 \\
    & 0.5000    & 0.2298    & 0.2622    & 0.1320    & 0.1372    & 0.1429    & 0.1116    & 0.0001    & 0.0022    & 4.6669\\
    & 1.0000    & 0.2002    & 0.2279    & 0.0903    & 0.0652    & 0.0680    & 0.0441    & 0.0002    & 0.0033    & 9.4382\\
    & 2.0000    & 0.1742    & 0.1985    & 0.0650    & 0.0307    & 0.0323    & 0.0179    & 0.0003    & 0.0057   & 19.0233\\
    % 5.0000    0.1450    0.1652    0.0461    0.0112    0.0119    0.0058    0.0005    0.0147   47.9025
 \hline
    & 0.0100    & 0.6986    & 0.7179    & 0.2714   & 29.3331   & 29.6279   & 19.9200    & 0.0001    & 0.0002    & 0.0831\\
    & 0.0200    & 0.6026    & 0.6199    & 0.2071   & 13.8366   & 13.9978    & 8.7139    & 0.0001    & 0.0002    & 0.1697\\
    & 0.0500    & 0.5026    & 0.5169    & 0.1452    & 5.1074    & 5.1724    & 2.8475    & 0.0001    & 0.0004    & 0.4440\\
    & 0.1000    & 0.4371    & 0.4496    & 0.1157    & 2.4004    & 2.4320    & 1.2555    & 0.0001    & 0.0006    & 0.9195\\
   14 & 0.2000    & 0.3807    & 0.3916    & 0.0894    & 1.1236    & 1.1389    & 0.5365    & 0.0001    & 0.0010    & 1.9072\\
    & 0.5000    & 0.3169    & 0.3260    & 0.0638    & 0.4101    & 0.4159    & 0.1757    & 0.0001    & 0.0022    & 4.9003\\
    & 1.0000    & 0.2762    & 0.2840    & 0.0497    & 0.1913    & 0.1941    & 0.0738    & 0.0002    & 0.0035    & 9.9515\\
    & 2.0000    & 0.2404    & 0.2472    & 0.0388    & 0.0889    & 0.0903    & 0.0302    & 0.0003    & 0.0055   & 20.1263\\
    % 5.0000    0.2001    0.2058    0.0280    0.0323    0.0328    0.0088    0.0005    0.0140   51.0985\\
\hline 
    & 0.0100    & 0.7044    & 0.7231    & 0.2236   & 33.5149   & 33.7061   & 18.1842    & 0.0001    & 0.0002    & 0.0791\\
    & 0.0200    & 0.6215    & 0.6373    & 0.1796   & 16.1934   & 16.3224    & 8.1126    & 0.0001    & 0.0002    & 0.1713\\
    & 0.0500    & 0.5164    & 0.5295    & 0.1367    & 5.9831    & 6.0581    & 2.8509    & 0.0001    & 0.0004    & 0.4384\\
    & 0.1000    & 0.4496    & 0.4610    & 0.1143    & 2.7692    & 2.8119    & 1.3265    & 0.0001    & 0.0006    & 0.8907\\
   15 & 0.2000    & 0.3924    & 0.4022    & 0.0948    & 1.2618    & 1.2839    & 0.5925    & 0.0001    & 0.0010    & 1.8308\\
    & 0.5000    & 0.3262    & 0.3345    & 0.0738    & 0.4410    & 0.4493    & 0.1936    & 0.0001    & 0.0022    & 4.7281\\
    & 1.0000    & 0.2841    & 0.2913    & 0.0561    & 0.1987    & 0.2024    & 0.0721    & 0.0002    & 0.0034    & 9.7408\\
    & 2.0000    & 0.2474    & 0.2536    & 0.0347    & 0.0905    & 0.0920    & 0.0200    & 0.0002    & 0.0055   & 19.9594\\
   % & 5.0000    & 0.2059    0.2111    0.0230    0.0324    0.0328    0.0046    0.0006    0.0145   50.8321\\
\hline
\end{tabular}
\label{bwkde1dexp3}}
\end{table}%\clearpage
%\section{conclusion}
%\label{bwkde1dconclusion} 
%More simpler expressions for $h_{cum}$ can be obtained with either symmetric PDF i.e. $k_3=0$ assumption or $\sigma =1$ condition or both. \\
%With assumption $k_3=0$
%\begin{equation}
%%\mbox{With $k_3=0$ assuption:}  \nonumber \\
%h_{cum} =  1.0592\sigma N^{-\frac{1}{5}}\left( 1 + 2.2559\frac{k_4}{\sigma^8} + \frac{k_4}{\sigma^{12}}\right) 
%\end{equation}
%%\mbox{With $\sigma=1$ condition:}  \nonumber \\
%With $\sigma=1$ condition:
%\begin{equation}
%h_{cum} = 1.0592 N^{-\frac{1}{5}}\left( 1 + 6.5625 k_3 + 2.2559 k_4 - k_3 - k_3k_4 + k_4\right)	
%\end{equation}
%%\mbox{With both $k_3=0$ assumption and $\sigma=1$ condition:}  \nonumber \\
%With both $k_3=0$ assumption and $\sigma=1$ condition:
%\begin{equation}
%h_{cum} = 1.0592 N^{-\frac{1}{5}}\left( 1 + 2.2559k_4 + k_4\right)	
%\end{equation}
%. The method by Raykar and Duraiswami [cite ] achieves comutation reduction for gaussian kernel in the \textit{solve-the-equation} based \textit{plug-in} methods by factorizing Gaussian through Taylor series.  
%Assuming change of variables, $z=\frac{x-s}{h}$, $s = hz + x$ and ${dz} = \frac{dx}{h}$ yields
%\begin{equation*}
%E\left[{\hat{f(x)}}\right] =  \int_{-\infty}^{\infty}{K(z)f(x-hz)dz}	
%\end{equation*}
%ICA is an established tool for both CA and BSS\citep{Cardoso98a,Comon2010,Duda2000}.
%\small
%\bibliographystyle{IEEEtran}
%\bibliography{Allref}
%%%\end{thebibliography}
%\end{document}
%\section{Multivariate Gram-Charlier Series for ExROT}
%\label{appndGC}
\section{Prerequisites for Multivariate ExROT}
\label{ndprereq}
The derivation for multivariate ExROT requires expressions for multivariate KDE, multivariate Taylor Series,  multivariate cumulants, multivariate Hermite polynomials, multivariate GCA Series and multivariate AMISE for bandwidth selection in KDE. Due to the reasons discussed in the Section \ref{introduction}, the article uses the expressions in  vector notations. The required pre-requisites are reported in this section and the multivariate AMISE is derived in the next section. 
%Conventional expressions use matrix derivatives for gradients of a multivariate PDF. Corresponding to that, the higher order derivatives require Tensor calculus. Often, the derived final expressions through Tensors require coordinatewise representations. Recently, there has been derived higher order cumulants  \citep{ndHermite02,SankhyaCumVec06} and multivariate Hermite polynomials \citep{ndHermite02,dnHermite96} using only elementary calculus of several variables, instead of the Tensor calculus. This is achieved by replacing conventional multivariate differentiations by repeated applications of the Kronecker product to vector differential operator. The derived expressions are also more elementary as using vector notations and more comprehensive as apparently more nearer to their counterparts in univariate. The same approach has been used in the next Section \ref{ndAMISE} to derive multivariate AMISE criteria in a vector notations. The remaining prerequisite expressions through Kronecker product are briefed in this section. 
\subsection{KDE for Multivariate PDF}
%The ExROT in multivariate KDE requires multivariate Gram-Charlier series, which has been derived in the previous section. 
\label{ndKDE}
The nonparametric estimation of a multivariate PDF requires a multivariate kernel. A multivariate kernel can be derived in two ways. 
%- either through product of univariate kernels or through selecting spherically symmetric multidimensional PDF as a kernel.
The most popular technique is to use a product kernel i.e. a d-dimensional kernel is achieved through the product of 1-dimensional kernels in each direction. The product kernel may have unequal spread or bandwidth in each direction. The other way is to select a spherically symmetric multidimensional PDF as the kernel, called spherical kernel. 
%Usually, the unknown data is whiten to have equal variance in each direction

\hspace{0.2 in}	 Let $\mathbf{x} = (x_1,x_2,\ldots,x_d)$ be a d-dimensional random vector. Let $\mathcal{K}(\mathbf{u})$ be the d-dimensional kernel. Then, the multivariate PDF of $\mathbf{x}$ with available  N realizations can be given as under:
\begin{align}
\label{eqndkde}
 \hat{f(\mathbf{x})} &= \frac{1}{N}\sum_{i=1}^{N}\frac{1}{\mbox{det}(\mathbf{H})}\mathcal{K}\left(\mathbf{H}^{-1}(\mathbf{x} - \mathbf{x}_{i})\right) = \frac{1}{N}\sum_{i=1}^{N}\mathcal{K}_{\mathbf{H}}(\mathbf{x} - \mathbf{x}_{i})
\end{align}
where, $\mathbf{H}$ is the bandwidth matrix, $\mbox{det}(\cdot)$ implies the determinant of a matrix and the symbol  $\mathcal{K}_{\mathbf{H}}(\cdot) = \frac{1}{\mbox{det}(\mathbf{H})}\mathcal{K}\left(\mathbf{H}^{-1}(\cdot)\right)$. In a correlated data, $\mathbf{H}$ is made proportional to the covariance matrix $\mathbf{C}_\mathbf{x}$ to compensate for the mutual correlations. Usually,  the kernel $\mathcal{K}(\mathbf{u})$ is a multiplicative kernel given by  $\mathcal{K}(\mathbf{u}) = \prod_{i=1}^{d}\mathcal{K}(u_i)$. Also, it is  symmetric, positive definite, zero mean and bounded function; usually a PDF. 

\hspace{0.2 in}%	For an uncorrelated data, 
Let $\mathcal{F}$ be the class of all symmetric, positive definite $d \times d$ matrices. Then, $\mathbf{H} \in \mathcal{M}$. A simplification can be achieved, if $\mathbf{H} \in \mathcal{D}$, where $\mathcal{D} \subseteq \mathcal{M}$ is a set of $d \times d$ diagonal matrices. More precisely, $\mathbf{H} = \mbox{diag}(h_1,h_2,\ldots,h_d)$. % i.e. unequal bandwidth in each direction. 
 The multivariate PDF  $\mathbf{x}$ with available  N realizations is now given as under:
\begin{align}
 \label{eqndkdemul}
 \hat{f(\mathbf{x})} &= \frac{1}{N}\sum_{i=1}^{N}\frac{1}{h_1h_2\ldots h_d}\mathcal{K}\left(\frac{x_1 - x_{i1}}{h_1}, \ldots,\frac{x_d - x_{id}}{h_d}\right) \\
 &= \frac{1}{N}\sum_{i=1}^{N}\left\{ \prod_{j=1}^{d}h_j^{-1}\mathcal{K}\left(\frac{x_j - x_{ij}}{h_j}\right) \right\}
\end{align} 
Further simplification is achieved  if $\mathbf{H} \in \mathcal{S}$, where $\mathcal{S} \subseteq \mathcal{D}$ is a set set of matrices with constant diagonal. More precisely, $\mathbf{H} = h\mathbf{I}_d$ i.e. equal bandwidth in all directions.
\subsection{Taylor Series of a multivariate function}
Let $\mathbf{x} = \left(X_1, X_2,..., X_d\right)'$ be a d-dimensional column vector and $f(\mathbf{x})$ be the function of several variables differentiable in each variable. 
The definition and application of Kronecker product on vector derivative operator $\mathbf{D}_{\mathbf{x}} = \left(\frac{\partial}{\partial x_1}, \frac{\partial}{\partial x_2}, \cdots, \frac{\partial}{\partial x_d}\right)'$  are briefed in Appendix \ref{KronDer}. Using them, the Taylor Series in vector notations for $f(\mathbf{x})$ by expanding it at origin, is given as under:
\begin{equation}
\label{ndTaylor}
f(\mathbf{x}) = \sum_{m=0}^{m=\infty}\frac{1}{m!}\mathbf{k}(m,d)'\mathbf{x}^{\otimes m}
\end{equation} 
where, $\mathbf{k}(m,d)$ is the vector of dimension $d^m \times 1$ and given in terms of the derivative vector $\mathbf{D}_{\mathbf{x}}$ as 
%=\left(\frac{\partial}{\partial x_1}, \frac{\partial}{\partial x_2}, \cdots, \frac{\partial}{\partial x_d}\right)'$  as
\begin{equation*}
%\label{ndcmd}
\mathbf{k}(m,d) = \left( \mathbf{D}_{\mathbf{x}}^{\otimes m} f(\mathbf{x})\right)\vline_{\mathbf{x}=\mathbf{0}}
\end{equation*}
\subsection{Moments, Cumulants and Hermite Polynomials}
Let $\mathbf{x} = \left(X_1, X_2,..., X_d\right)'$ be a d-dimensional random vector, $f(\mathbf{x})$ be its joint PDF differentiable in each variable. Also, let for $\bs{\lambda} = (\lambda_1, \lambda_2, \ldots, \lambda_d)'$,  $\bs{\lambda} \in \mathbb{R}^d$ and $f(\mathbf{x})$ - $\mathcal{F}(\bs{\lambda})$ be the Characteristic function,  $\mathbf{M}(\bs{\lambda})$ be the Moment Generating Function (MGF), $\mathcal{C}(\bs{\lambda})$ be the other form of  Characteristic function through natural log of $\mathcal{F}(\bs{\lambda})$, $\mathbf{C}(\bs{\lambda})$ be the Cumulant Generating Function (CGF). Then, assuming $\mathbf{M}(\bs{\lambda})$ and $\mathcal{F}(\bs{\lambda})$ have been expanded using Taylor Series,
\begin{equation}
\label{momkd}
\mathbf{m}(k,d) = \int_{-\infty}^{\infty}\mathbf{x}^{\otimes k}f(\mathbf{x})d\mathbf{x} =\mathbf{D}_{\bs{\lambda}}^{\otimes k} \mathbf{M}(\bs{\lambda})\vline_{\bs{\lambda}=\mathbf{0}} = (-i)^k \mathbf{D}_{\bs{\lambda}}^{\otimes k} \mathcal{F}_{\mathbf{x}}(\bs{\lambda})\vline_{\bs{\lambda}=\mathbf{0}} 
\end{equation}
Assuming $\mathbf{C}(\bs{\lambda})$ and $\mathcal{C}(\bs{\lambda})$ have been expanded using Taylor Series, 
\begin{equation}
\label{cumkd}
\mathbf{c}(k,d) = \mathbf{D}_{\bs{\lambda}}^{\otimes k} \mathbf{C}(\bs{\lambda})\vline_{\bs{\lambda}=\mathbf{0}} = (-i)^k \mathbf{D}_{\bs{\lambda}}^{\otimes k} \mathcal{C}_{\mathbf{x}}(\bs{\lambda})\vline_{\bs{\lambda}=\mathbf{0}} 
\end{equation}
where, $\mathbf{C}(\bs{\lambda}) = \ln \mathbf{M}(\bs{\lambda})$ and $\mathcal{C}(\bs{\lambda})= ln \mathcal{F}(\bs{\lambda})$. 
The details on the moments, cumulants and their relationships can be found in \citep{ndGGCarxivDharmani}.

\hspace{0.2 in}	The multivariate Hermite polynomials are defined as under in Equation \eqref{hermitend}.
\begin{align}
\label{hermitend}
\mathbf{H}_k(\mathbf{x}; \mathbf{0},\mathbf{C}_{\mathbf{x}}) &= [G(\mathbf{x};\mathbf{0},\mathbf{C}_{\mathbf{x}})]^{-1} (-1)^{k} \left( \mathbf{C}_{\mathbf{x}} \mathbf{D}_{\mathbf{x}} \right)^{\otimes k} G(\mathbf{x};\mathbf{0},\mathbf{C}_{\mathbf{x}}) \\
\mbox{where, } G(\mathbf{x};\mathbf{0},\mathbf{C}_{\mathbf{x}}) &= \left|\mathbf{C}_{\mathbf{x}} \right|^{-1/2}(2\pi)^{-d/2}\exp\left(\mathbf{x}' \mathbf{C}_{\mathbf{x}}^{-1} \mathbf{x} \right) = \left|\mathbf{C}_{\mathbf{x}} \right|^{-1/2}(2\pi)^{-d/2}\exp\left( \left(\mbox{vec}\mathbf{C}_{\mathbf{x}}^{-1}\right)' \mathbf{x}^{\otimes 2} \right)
%\mathbf{H}_k(\mathbf{x}; \mathbf{0},\mathbf{C}_{\mathbf{x}}) &= [G(\mathbf{x};\mathbf{0},\mathbf{C}_{\mathbf{x}})]^{-1} (-1)^{k} (\mathbf{C}_{\mathbf{x}} \mathbf{D}_x))^{\otimes k} G(\mathbf{x};\mathbf{0},\mathbf{C}_{\mathbf{x}})
%\mbox{where, } G(\mathbf{x};\mathbf{0},\mathbf{C}_{\mathbf{x}}) &= \left|\\mathbf{C}_{\mathbf{x}} \right|^{-1/2}(2\pi)^{-d/2}\exp\left(\mathbf{x}'\left(\\mathbf{C}_{\mathbf{x}} \right)^{-1}\mathbf{x} \right)
\end{align}
The d-variate Hermite polynomials, derived  by \citet[Proposition 6]{ndHermite02}, are given by the following recursion  rule:
\begin{align}
\mathbf{H}_0(\mathbf{x}) &= 1 ; \mathbf{H}_1(\mathbf{x}) =  \mathbf{x} \\
\mbox{For } n > 1 & \nonumber \\
\label{eqndHermite}
 \mathbf{H}_n(\mathbf{x}_{(1:n)}) & =  \mathbf{H}_{n-1}(\mathbf{x}{(1:n-1)}) \otimes \mathbf{x}  \nonumber \\ & - \sum_{j=1}^{n-1}\mathbf{K}^{-1}_{\mathcal{P}(n,j)\rightarrow (1,2)}(d_{(1:n)}) \times \mathbf{c}_{n,j}(2,d) \otimes \mathbf{H}_{n-2}(\mathbf{x}_{(1:j-1,j+1:n-1)})
 \end{align}
where, the permutation $\mathcal{P}(n,j)\rightarrow (1,2)$ is putting the $n^{th}$ element to the first and $j^{th}$ to the second place,  keeping  the other elements unchanged. More precisely,
\begin{align*}
\mathbf{K}^{-1}_{\mathcal{P}(n,j)\rightarrow (1,2)}(d_{(1:n)}) &= \left( \mathbf{K}_{\mathcal{P}j+1\rightarrow 2}(d_n, d_{(1:n-1)}) \mathbf{K}_{\mathcal{P}n\rightarrow 1}(d_n, d_{(1:n-1)}) \right)^{-1} \\
& = \mathbf{K}^{-1}_{\mathcal{P}n\rightarrow 1}(d_n, d_{(1:n-1)}) \mathbf{K}^{-1}_{\mathcal{P}j+1\rightarrow 2}(d_n, d_{(1:n-1)}) 
\end{align*}
\subsection[A Compact Derivation for multivariate GGC Series]{A Compact Derivation for Multivariate GGC Series and GCA Series} %Generalized Gram-Charlier
\label{compactGGC}
A full derivation for multivariate GGC and corresponding GCA can be found in \cite{ndGGCarxivDharmani}. 
This is a compact derivation for Generalized Gram-Charlier Series. The detailed proof can be found in the article. 
Let $\mathcal{F}_{\mathbf{x}}$ be the characteristic function of an unknown multivariate PDF $f(\mathbf{x})$. Let $\Psi(\mathbf{x})$ be the characteristic function of a reference multivariate PDF $\psi(\mathbf{x})$. 
 Also, let $\mathbf{c}(k,d)$ be the $k^{th}$ cumulant of d-dimensional $f(\mathbf{x})$ and  $\mathbf{c}_r(k,d)$ be the $k^{th}$ cumulant of d-dimensional $\psi(\mathbf{x})$. The difference of the cumulant vector is given by $\bs{\delta}(k,d) = \mathbf{c}(k,d) - \mathbf{c}_r(k,d)$. Instead of the detailed proof, a compact derivation of GGC follows here.

\hspace{0.2 in} Let $\bs{\lambda} = (\lambda_1, \lambda_2, \ldots, \lambda_d)^T$, $\lambda \in \mathbb{R}^d$; $\otimes$ be the Kronecker product operator and $\mathsf{F}$ is the Fourier transform operator. Then, 
%$f(\lambda) \in \mathbb{R}^{m_1}$  and $g(\lambda) \in \mathbb{R}^{m_2}$. Then
\begin{align}
\mathcal{F}_{\mathbf{x}}(\bs{\lambda}) & = \exp\left[ \sum_{k=1}^{\infty}\mathbf{c}(k,d)  \frac{(i\bs{\lambda})^{\otimes k}}{k!}\right] & (\mbox{ $\because$ definition })\\
& = \exp\left[ \sum_{k=1}^{\infty}\delta(k,d)  \frac{(i\bs{\lambda})^{\otimes k}}{k!}\right]\exp\left[ \sum_{k=1}^{\infty}\mathbf{c}_r(k,d)  \frac{(i\bs{\lambda})^{\otimes k}}{k!}\right] &\\
& = \left[ \sum_{k=0}^{\infty}\alpha(k,d)  \frac{(i\bs{\lambda})^{\otimes k}}{k!}\right]\mathsf{F}(\psi(\mathbf{x})) 
\end{align}
%whre, $\mathsf{F}$ is the Fourier transform oprator and $\otimes$ is the Kronecker product operator.
Taking inverse Fourier transform of the above equation brings 
\begin{align}
f(\mathbf{x}) &= \sum_{k=0}^{\infty}\alpha(k,d)  \frac{(-1)^k}{k!}\delta^{(k)}(\mathbf{x}) * \psi(\mathbf{x}) \\
\label{GGC}
f(\mathbf{x}) &= \sum_{k=0}^{\infty}\alpha(k,d)  \frac{(-1)^k}{k!}\psi^{(k)}(\mathbf{x}) 
\end{align}
where, $*$ indicates convolution. Thus, the GGC is obtained in a compact way.

\hspace{0.2 in}	The $\alpha(k,d)$ can be derived using its equality in terms of $\bs{\delta}(k,d)$ as:
\begin{align*}
\sum_{k=0}^{\infty}\bs{\alpha}(k,d)  \frac{(i\bs{\lambda})^{\otimes k}}{k!} = \exp\left[ \sum_{k=1}^{\infty}\bs{\delta}(k,d)  \frac{(i\bs{\lambda})^{\otimes k}}{k!}\right]
\end{align*}
This results into the following relations:
\begin{align*}
\bs{\alpha}(0,d) =& 1 \\
\bs{\alpha}(1,d) =& \bs{\delta}(1,d) \\
\bs{\alpha}(2,d) =& \bs{\delta}(2,d) + \bs{\delta}(1,d)^{\otimes 2} \\ 
\bs{\alpha}(3,d)=& \bs{\delta}(3,d) + 3\bs{\delta}(2,d)\otimes \bs{\delta}(1,d) + \bs{\delta}(1,d)^{\otimes 3} \\
\bs{\alpha}(4,d) =&  \bs{\delta}(4,d) + 4\bs{\delta}(3,d)\otimes \bs{\delta}(1,d) + 3\bs{\delta}(2,d)^{\otimes 2} + 6 \bs{\delta}(2,d)\otimes \bs{\delta}(1,d)^{\otimes 2} + \bs{\delta}(1,d)^{\otimes 4}	\\
\bs{\alpha}(5,d) =& \bs{\delta}(5,d) + 5\bs{\delta}(4,d)\otimes \bs{\delta}(1,d)+ 10\bs{\delta}(3,d)\otimes \bs{\delta}(2,d) + 10\bs{\delta}(3,d)\otimes \bs{\delta}(1,d)^{\otimes 2} \\
& + 15\bs{\delta}(2,d)^{\otimes 2}\otimes \bs{\delta}(1,d) + 10\bs{\delta}(2,d)\otimes \bs{\delta}(1,d)^{\otimes 3} 
  + \bs{\delta}(1,d)^{\otimes 5} \\
\bs{\alpha}(6,d) =& \bs{\delta}(6,d) + 6\bs{\delta}(5,d)\otimes \bs{\delta}(1,d) + 15 \bs{\delta}(4,d) \otimes \bs{\delta}(2,d) + 15 \bs{\delta}(4,d) \otimes \bs{\delta}(1,d)^{\otimes 2} + 10 \bs{\delta}(3,d)^{\otimes 2} \\
&+ 60\bs{\delta}(3,d)\otimes \bs{\delta}(2,d)\otimes \bs{\delta}(1,d) 
+ 20\bs{\delta}(3,d)\otimes \bs{\delta}(1,d)^{\otimes 2} + 15\bs{\delta}(2,d)^{\otimes 3} \\
&+ 45\bs{\delta}(2,d)^{\otimes 2}\otimes \bs{\delta}(1,d)^{\otimes 2}  + 15\bs{\delta}(2,d)\otimes \bs{\delta}(1,d)^{\otimes 4} + \bs{\delta}(1,d)^{\otimes 6}  
\end{align*}
%The above equations have exact resemblance with those expressing moments in terms of the cumulants in Section \ref{RelCumMom}. 
The multivariate GGC, for specific Gaussian PDF reference and matching first and second order moments, can be obtained by taking $\bs{\delta}(1,d) = 0$, $\bs{\delta}(2,d)=0$ and $\mathbf{c}_r(k,d) = \mathbf{0}$.  %i.e. mean, unit variance in all directions and same as that of the unknown PDF.
%\subsection{Multivariate Gram-Charlier A Series}
The GGC with respect to a Gaussian PDF is identified as GCA Series, given as under: 
\begin{align} 
\label{GCA}
f(\mathbf{x}) & = G(\mathbf{x}) -   \frac{\mathbf{c}(3,d)'}{3!} G^{(3)}(\mathbf{x}) +  \frac{\mathbf{c}(4,d)'}{4!}G^{(4)}(\mathbf{x}) - \frac{\mathbf{c}(5,d)'}{5!} G^{(5)}(\mathbf{x}) \nonumber \\
& + \frac{\mathbf{c}(6,d)' + 10 \mathbf{c}(3,d)^{\otimes 2'}}{6!} G^{(6)}(\mathbf{x}) + \ldots
\end{align}
\section{AMISE in vector notations for bandwidth selection multivariate KDE}
\label{ndAMISE}
The derivation for bandwidth parameter selection, satisfying AMISE between the estimated multivariate PDF and the actual multivariate PDF, %criteria to estimate the quality of PDF estimation in multivariate 
can be found in \citep{wand1994kernel}. For the reasons described before, it is re-derived here in terms of vector notations using Kronecker product and differential operator as a column vector $\mathbf{D}_{\mathbf{x}} = \left(\frac{\partial}{\partial x_1}, \frac{\partial}{\partial x_2}, \cdots, \frac{\partial}{\partial x_d}\right)'$.  %The brief is provided here. The crietria is given as: 
 \begin{align}
 \label{eqndmise}
 \mbox{MISE}( f(\mathbf{x}),{\hat{f(\mathbf{x})}} ) &=  \int_{-\infty}^{\infty}{\mbox{Bias}^2({\hat{f(\mathbf{x})}})d\mathbf{x}} + \int_{-\infty}^{\infty}{\mbox{Var}({\hat{f(\mathbf{x})}})d\mathbf{x}} 
 \end{align} 
%From, Equation \eqref{ndTaylor}, the Taylor series expansion of a multivariate function $f(\cdot)$ near $\mathbf{x}$ approximated upto the second order derivative is given by 
%%\begin{align}
%%\label{oldndAMISE}
%%f(\mathbf{x} + \lambda) = f(\mathbf{x}) + \lambda'\nabla_f(\mathbf{x}) + \frac{1}{2}\lambda'\mathcal{H}_f(\mathbf{x})\lambda + o(\lambda'\lambda) 
%%\end{align}
%%where, $\nabla(\cdot)$ is a gradient matrix and $\mathcal{H}(\cdot)$ is a Hessian matrix. Let the derivative operator be  $\mathbf{D}_\mathbf{x} = (\frac{\partial}{\partial x_1}, \frac{\partial}{\partial x_2},\ldots, \frac{\partial}{\partial x_d})'$. Then, instead of this conventional notations, vectorization of the Jacobian through Kronecker product with derivative opertor 
%%%as in equation \eqref{ndTaylor} 
%%simplifies the expression as under:
%\begin{align}
%f(\mathbf{x} + \lambda) 
%%&= f(\mathbf{x}) + (\mathbf{D}^{\otimes }_{\mathbf{x}}f(\mathbf{x}))'\lambda + \frac{1}{2}( \mathbf{D}^{\otimes 2}_{\mathbf{x}}f(\mathbf{x}))'\lambda^{\otimes 2} + o(\lambda'\lambda) \\
%&= f(\mathbf{x}) + \mathbf{k}(1,d)'\lambda + \frac{1}{2}\mathbf{k}(2,d)'\lambda^{\otimes 2} + O(\lambda'\lambda) \\
%\mbox{where, } \mathbf{k}(i,d) &= \mathbf{D}^{\otimes i}_{\mathbf{x}}f(\mathbf{x})\left. \right|_{\mathbf{x} = \mathbf{0}} \nonumber 
%\end{align}
The bias and variance estimations, using the Taylor Series expansion in Equation \eqref{ndTaylor}, are derived as under.  
\begin{align} 
E\{\hat{f(\mathbf{x})}\} &= \frac{1}{N}\sum_{i=1}^{N}{E\left\{\mathcal{K}_{\mathbf{H}} (\mathbf{x}-\mathbf{x}_i) \right\} }	 \nonumber \\
&= \int_{-\infty}^{\infty}{ \mathcal{K}_{\mathbf{H}}(\mathbf{u}-\mathbf{x}) f(\mathbf{u})d\mathbf{u} }	 \nonumber \\
&=   \int_{-\infty}^{\infty}{ \mathcal{K}(\mathbf{z})f(\mathbf{x}+\mathbf{Hz})d\mathbf{z} } \mbox{   ( $\because$ substituting $z=\mathbf{u}-\mathbf{x}$ ) }  \nonumber \\
&= \int_{-\infty}^{\infty} \mathcal{K}(\mathbf{z})\left( f(\mathbf{x}) + \mathbf{k}(1,d)' (\mathbf{Hz}) + \frac{1}{2}\mathbf{k}(2,d)'(\mathbf{Hz})^{\otimes 2} + O( \mbox{tr}(\mathbf{H}^{\otimes 2}) ) \right) d\mathbf{z} \nonumber 
%&= f(\mathbf{x}) + \frac{1}{2}\mu_2(\mathcal{K})\mbox{tr}(\mathbf{H}'\mathbf{D}^{\otimes 2}_f(\mathbf{x})\mathbf{H}) + O(\mathbf{H}'\mathbf{H})	 \nonumber \\
\end{align}
where, $\mathbf{k}(i,d) = \mathbf{D}^{\otimes i}_{\mathbf{x}}f(\mathbf{x})\left. \right|_{\mathbf{x} = \mathbf{x}}$ \\
%\hspace{0.2 in} 
Assuming the kernel with symmetric and bounded PDF the following properties are satisfied: 
\begin{align}
\int_{-\infty}^{\infty} \mathbf{u}\mathcal{K}(\mathbf{u})d\mathbf{u} &= 0 \\
\int_{-\infty}^{\infty} \mathbf{u}\otimes \mathbf{u} \mathcal{K}(\mathbf{u})d\mathbf{u} &= \mathbf{m}_{\mathcal{K}}(2,d) = \mu_2(\mathcal{K})\mbox{vec}(\mathbf{I}_{(d \times d)}) \\
& = \mu_2(\mathcal{K}) \bs{\delta}_2
\mbox{where, } \bs{\delta}_2 = \mbox{vec}(\mathbf{I}_{(d \times d)}) 
%(\bs\delta_k)_i = 
%\begin{cases} 1 \mbox{ ; for } i = 1+ \mathbf{r}(d^{k-1}+1), \mathbf{r}= 0:(d-1) \\
%0 \mbox{ ; otherwise}
%\end{cases} \nonumber
\end{align}
Here, $\mathbf{m}_{\mathcal{K}}(2,d)$ is the second order moment vector of the d-variate kernel with $d^2$ components and $\bs{\delta}_2 $ is a vector of size $(d^2 \times 1)$ with only selected d values being 1.  The second order moment of each component is constant with value  $\mu_2(\mathcal{K})$ and all cross-moments are zero.   %  = \int{z^2\mathcal{K}^2(z)dz}$ %Also, $\mathbf{1}_{(d^2 \times 1)} $ is a column vector of ones with $d^2$ elements. \\
%\hspace{0.2 in} 
Using these properties, the bias can be written as: 
\begin{align}
\label{eqndbias}
\Rightarrow \mbox{Bias}({\hat{f(\mathbf{x})}}) = E\{\hat{f(\mathbf{x})}\} - f(\mathbf{x})  &\approx 
 \frac{1}{2}  \mathbf{k}(2,d)'\left( \mathbf{H}^{\otimes 2}\mathbf{m}_{\mathcal{K}}(2,d) \right) %+ O(\mathbf{H}^{\otimes 2})
\end{align}
Similarly, the variance of the estimation can be approximated as under:
\begin{align}
\mbox{Var}({\hat{f}(\mathbf{x})}) &= \mbox{Var}\left( \frac{1}{N}\sum_{i=1}^{N}\mathcal{K}_{\mathbf{H}}( \mathbf{x}-\mathbf{x}_i) \right)  &\nonumber \\
&= \frac{1}{N\mbox{det}(\mathbf{H})}\int_{-\infty}^{\infty}{ \mathcal{K}^2(z) f(\mathbf{x}+ \mathbf{Hz})d\mathbf{z} }  
 - \frac{1}{N}E^2 \left\{ \hat{f}(\mathbf{x})\right\} \nonumber\\
&= \frac{1}{N\mbox{det}(\mathbf{H})}\int_{-\infty}^{\infty}\mathcal{K}^2(z) \left( f(\mathbf{x}) + \mathbf{k}(1,d)'\mathbf{Hz} + \mathbf{k}(2,d)'(\mathbf{Hz})^{\otimes 2}  + O(\mbox{tr}(\mathbf{H}^{\otimes 2})) \right) d\mathbf{z}  \nonumber \\
 & \mbox{ } - \frac{1}{N}E^2 \left\{ \hat{f}(\mathbf{x})\right\} \nonumber\\
\label{eqndvar}
\Rightarrow \mbox{Var}( {\hat{f}(\mathbf{x}) }) &\approx \frac{1}{N\mbox{det}(\mathbf{H})}f(\mathbf{x})\int_{-\infty}^{\infty}{\mathcal{K}^2(\mathbf{z})d\mathbf{z}}  \mbox{    ($\because$ assuming large $N$, small $h$ ) }% and given kernel properties
\end{align}
Combining equations \eqref{eqndmise}, \eqref{eqndbias} and \eqref{eqndvar}; we get:
\begin{align}
\mbox{MISE}({\hat{f}(\mathbf{x})}) & = \frac{1}{4} %(\mu_2(\mathcal{K}))^2 
\int_{-\infty}^{\infty} \left( \mathbf{k}(2,d)' \mathbf{H}^{\otimes 2}\mathbf{m}_\mathcal{K}(2,d) \right)^2 d\mathbf{z} +  \frac{R(\mathcal{K})}{N\mbox{det}(\mathbf{H})} + O\left(\mbox{tr}\left(\mathbf{H}^{\otimes 4}\right)\right) + O\left(\frac{\mbox{det}(\mathbf{H})}{N}\right) \nonumber 
\end{align}
where, %$\mu_2(\mathcal{K}) = \int{z^2\mathcal{K}^2(z)dz}$ and 
$R(\mathcal{K}) = \int_{-\infty}^{\infty}{\mathcal{K}^2(z)dz}$. 
An asymptotic large sample approximation AMISE is obtained, assuming $\lim_{N\rightarrow\infty}{\mbox{det}(\mathbf{H})} = 0$ and $\lim_{N\rightarrow\infty}{N\mbox{det}(\mathbf{H})} = \infty$ i.e. $\mbox{det}(\mathbf{H})$ reduces to 0 at a rate slower than $1/N$.
\begin{align}
\mbox{AMISE}({\hat{f}(\mathbf{x})}) &= \frac{1}{4} \int_{-\infty}^{\infty} \left( \mathbf{k}(2,d)' \mathbf{H}^{\otimes 2}\mathbf{m}_\mathcal{K}(2,d) \right)^2 d\mathbf{z} +  \frac{1}{N\mbox{det}(\mathbf{H})}R(\mathcal{K}) \nonumber \\
\label{eqndmisef0}
& = \frac{1}{4}\mu_2^2(\mathcal{K}) \int_{-\infty}^{\infty} \left( \mathbf{k}(2,d)' \left( \mathbf{H}^{\otimes 2} \bs{\delta}_2 \right) \right)^2  d\mathbf{z} +  \frac{1}{N\mbox{det}(\mathbf{H})}R(\mathcal{K}) \\
%\mbox{Now, }
%\left( \mathbf{k}(2,d)' \left( \mathbf{H}^{\otimes 2} \bs{\delta}_2 \right) \right)^2
%%& = \left( \left( \mathbf{k}(2,d) \mbox{ o } \bs{\delta}_2 \right) ' \left( \mathbf{H}^{\otimes 2} \bs{\delta}_2 \right) \right)^2  \mbox{ }(\because \bs{\delta}_k  \mbox{ is an indicator function)\nonumber \\
%&= \left( \mathbf{k}(2,d)' \left( \mathbf{H}^{\otimes 2} \bs{\delta}_2 \right) \right) \left( \mathbf{k}(2,d)' \left( \mathbf{H}^{\otimes 2} \bs{\delta}_2 \right) \right)' \nonumber \\
%\mbox{As } \mbox{ vec }(ab') = b \otimes a \mbox{ and }\mathbf{x'Ax}= (\mbox{vec}\mathbf{A})'\mathbf{x}^{\otimes 2}, \nonumber \\ 
%\left( \mathbf{k}(2,d)' \left( \mathbf{H}^{\otimes 2} \bs{\delta}_2 \right) \right)^2 &=  \left( \left( \mathbf{H}^{\otimes 2} \bs{\delta}_2 \right)^{\otimes 2} \right)' \left( \mathbf{k}(2,d)  \right)^{\otimes 2}
%\mbox{Now, as }  \bs{\delta}_k & \mbox{ is an indicator function, }\\
\mbox{Now, }
\left( \mathbf{k}(2,d)' \left( \mathbf{H}^{\otimes 2} \bs{\delta}_2 \right) \right)^2
& = \left( \left( \mathbf{k}(2,d) \mbox{ o } \bs{\delta}_2 \right) ' \left( \mathbf{H}^{\otimes 2} \bs{\delta}_2 \right) \right)^2   \nonumber \\ %\mbox{ }(\because \bs{\delta}_k  \mbox{ is an indicator function })
%&= \left( \left( \mathbf{k}(2,d) \mbox{ o } \bs{\delta}_2 \right)' \left( \mathbf{H}^{\otimes 2} \bs{\delta}_2 \right) \right) \left( \left( \mathbf{k}(2,d) \mbox{ o } \bs{\delta}_2 \right)' \left( \mathbf{H}^{\otimes 2} \bs{\delta}_2 \right) \right)' \nonumber \\
%\mbox{As } \mbox{ vec }(ab') = b \otimes a  \mbox{ and }  & \mathbf{x'Ax}= (\mbox{vec}\mathbf{A})'\mathbf{x}^{\otimes 2}, \nonumber \\ 
%\left( \mathbf{k}(2,d)' \left( \mathbf{H}^{\otimes 2} \bs{\delta}_2 \right) \right)^2 
&=  \left( \left( \mathbf{H}^{\otimes 2} \bs{\delta}_2 \right)^{\otimes 2} \right)' \left( \mathbf{k}(2,d) \mbox{ o } \bs{\delta}_2 \right)^{\otimes 2} \mbox{ }(\because \mathbf{x'Ax}= (\mbox{vec}\mathbf{A})'\mathbf{x}^{\otimes 2}) \nonumber\\
% \mbox{  }( \because \mathbf{x'Ax}= (\mbox{vec}\mathbf{A})'\mathbf{x}^{\otimes 2}; \mbox{ vec }(ab') = b \otimes a ) 
\label{eqndmisef}
\Rightarrow \mbox{AMISE}({\hat{f}(\mathbf{x})}) &= \frac{1}{4}\mu_2^2(\mathcal{K}) \left( \mathbf{H}^{\otimes 4} \bs{\delta}_2^{\otimes 2} \right)' \mathbf{R}(f''(\mathbf{x})) +  \frac{1}{N\mbox{det}(\mathbf{H})}R(\mathcal{K})   \\
\mbox{where, }\mathbf{R}(f''(\mathbf{x})) &=  \int_{-\infty}^{\infty} \left( \mathbf{k}(2,d) \mbox{ o } \bs{\delta}_2 \right)^{\otimes 2}  d\mathbf{z} \nonumber
%& = \frac{1}{4}\mu_2^2(\mathcal{K}) \left( \int \mathbf{k}^2(2,d)  d\mathbf{z}\right)' \left( \mathbf{H}^{\otimes 2} \mathbf{1}_2 \right)^2 +  \frac{1}{N\mbox{det}(\mathbf{H})}R(\mathcal{K}) \\
%& = \frac{1}{4}\mu_2^2(\mathcal{K}) \mathbf{R}(\mathbf{k}(2,d))' \left( \mathbf{H}^{\otimes 2} \mathbf{1}_2 \right)^2 %\Psi_{\mathcal{F}}  +  \frac{1}{N \mbox{det}(\mathbf{H})} R(\mathcal{K}) 
\end{align}
%where, $\mathbf{R}(f''(\mathbf{x}))$ is the roughness vector of $f''(\mathbf{x})$ with dimension $d^4 \times 1$. 
The Equation \eqref{eqndmisef} interprets that a small $\mbox{det}(\mathbf{H})$ increases estimation variance, whereas, a larger $\mbox{det}(\mathbf{H})$ increases estimation bias. 
Also, $\mathbf{R}(f''(\mathbf{x}))$ is the roughness vector of $f''(\mathbf{x})$ with dimension $d^4 \times 1$ and the corresponding total roughness is given by,  
\begin{align}
\label{eqndrf2}
R(f''(\mathbf{x})) &= \sum_{i=1}^{d^4} [ \mathbf{R}(f''(\mathbf{x}) ]_i = \sum_{i=1}^{d^4} \left[ \int_{-\infty}^{\infty} \left( \mathbf{k}(2,d) \mbox{ o } \bs{\delta}_2 \right)^{\otimes 2}  d\mathbf{z} \right]_i   \\
& =	 \int_{-\infty}^{\infty} \left( \sum_{i=1}^{d}   \frac{\partial^2 f(\mathbf{x})}{\partial x_i^2} \right)^2 \mbox{ }(\because \bs{\delta}_2 \mbox{ has only d number of ones.}) \nonumber \\
&= \int_{-\infty}^{\infty} \left[ \left(  \nabla^2 f(\mathbf{x}) \right)^{2}\right]	d\mathbf{x} \nonumber
\end{align}
where, the indicator function $\bs{\delta}_2$ assures that only few partial derivatives are non-zero. 
%So, more precisely it is:
%\begin{align}
%R(f''(\mathbf{x})) &= \sum_{i=1}^{d^4} [ \mathbf{R}(f''(\mathbf{x})]_i 
% & = \int_{-\infty}^{\infty} \left(  \nabla^2 f(\mathbf{x}) \right)^2	d\mathbf{x} =	 \int_{-\infty}^{\infty} \left( \sum_{i=1}^{d}   \frac{\partial^2 f(\mathbf{x})}{\partial x_i^2} \right)^2
%\end{align}  
%where, % $ \Psi_{\mathcal{F}} = 
%$ \mathbf(R)(\mathbf{k}(2,d)) = \int \left( \mathbf{k}(2,d)\right)^2  d\mathbf{z} = \int \left( \mathbf{D}^{\otimes 2}f(\mathbf{x})\right)^2  d\mathbf{z}$ is the $d^2 \times 1$ column vector and the roughness of $f''(\mathbf{x})$ is simply, $R(\mathbf{k}(2,d)) = \sum_{j=1}^{d^2} \mathbf{R}_j(\mathbf{k}(2,d))$.
%The Equation \eqref{eqndmisef} interprets that a small $\mbox{det}(\mathbf{H})$ increases estimation variance, whereas, a larger $\mbox{det}(\mathbf{H})$ increases estimation bias.

\hspace{0.2 in}		To simplify the bandwidth estimation, let us assume $\mathbf{H} \in \mathcal{D}$ and $\mathcal{D}$ is the set of all diagonal matrices. That is,  $\mathbf{H} = \mbox{diag}(h_1, h_2, \ldots, h_d) = \mbox{diag} (\mathbf{h})$. Corresponding to that, $$ \mathbf{H}^{\otimes 2} \bs{\delta}_2  = \mathbf{h}^{\otimes 2} \mbox{ o } \bs{\delta}_2 = \mathbf{h}_{\bs{\delta}}^{\otimes 2} \mbox{ (symbolically)}$$ where, o denotes hadamard product.
% and  $\mathbf{h}_{\bs{\delta}}$ is the alloted symbol.  
%= \left( \mathbf{h}^2\right)^{\otimes 2}$$. 
Based on this simplification,  the   AMISE is given by,
\begin{align}
\label{eqnddamise}
%\mbox{AMISE}({\hat{f}(\mathbf{x})}) &= \frac{1}{4}\mu_2^2(\mathcal{K})\int_{-\infty}^{\infty} \left( \mathbf{k}(2,d)' \mathbf{h}_{\bs{\delta}}^{\otimes 2} \right)^2 d\mathbf{z} +  \frac{1}{N\prod_{i=1}^{d}h_i } R(\mathcal{K}) \\
%%\end{align}
%%\begin{align}
%\left( \mathbf{k}(2,d)' \mathbf{h}_{\bs{\delta}}^{\otimes 2} \right)^2
%&= \left( \mathbf{k}(2,d)' \mathbf{h}_{\bs{delta}}^{\otimes 2} \right) \left( \mathbf{k}(2,d)' \mathbf{h}_{\bs{\delta}}^{\otimes 2} \right)' \nonumber \\
%&=  \left( \mathbf{h}_{\bs{delta}}^{\otimes 4} \right)' \mathbf{k}(2,d)_{\otimes 2} \mbox{  }( \because \mathbf{x'Ax}= \mbox{vec}\mathbf{A}'\mathbf{x}^{\otimes 2} )  \\
%%&= \left( \sum_{i=1}^{d^2} \left( \mathbf{k}(2,d) \mbox{ o } \mathbf{h}_{\bs{delta}}^{\otimes 2} \right)_i \right)^2
%%&=  \left( \sum_{j=1}^{d^2} \sum_{i=1}^{d^2} \left( \mathbf{k}(2,d) \mbox{ o } \mathbf{h}_{\bs{delta}}^{\otimes 2} \right)_i \right)^2
%\Rightarrow \mbox{AMISE}({\hat{f}(\mathbf{x})}) &= \frac{ \left( \mathbf{h}^{\otimes 4}\right)' }{4}\mu_2^2(\mathcal{K})  \int_{-\infty}^{\infty} \left( \mathbf{k}(2,d) \mbox{ o }  \bs{\delta}  \right)^{\otimes 2} d\mathbf{z} +  \frac{1}{N\prod_{i=1}^{d}h_i}R(\mathcal{K})  \nonumber \\
\mbox{AMISE}({\hat{f}(\mathbf{x})}) &= \frac{\mu_2^2(\mathcal{K})}{4}\left( \mathbf{h}^{\otimes 4} \mbox{ o }\bs{\delta}_2^{\otimes 2}\right)' \mathbf{R}(f''(\mathbf{x}))  +  \frac{1}{N\prod_{i=1}^{d}h_i}R(\mathcal{K}) 
\end{align}
%where, $R(f''(\mathbf{x}))$ is the roughness of $f''(\mathbf{x})$ and it is given by,  
%\begin{align}
% R(f''(\mathbf{x})) &= \int_{-\infty}^{\infty} \left( \mathbf{k}(2,d)'\left( \mathbf{I}_d^{\otimes 2}\bs{\delta}_2 \right) \right)^2  d\mathbf{z} = \int_{-\infty}^{\infty} \left( \mathbf{k}(2,d)'\bs{\delta}_2 \right)^2  d\mathbf{z}  \\
% & =  \int_{-\infty}^{\infty} \left( \sum_{i=1}^{d^2} ( \mathbf{k}(2,d) \mbox{ o } \bs{\delta} )_i \right)^2  d\mathbf{z} = \int_{-\infty}^{\infty}  \sum_{j=1}^{d^2} \sum_{i=1}^{d^2} \left[ (\mathbf{k}(2,d) \mbox{ o }\bs{\delta}) (\mathbf{k}(2,d)\mbox{ o }\bs{\delta})'\right]_{i,j}   d\mathbf{z} \\
% & = \int_{-\infty}^{\infty} \left(  \nabla^2 f(\mathbf{x}) \right)^2	d\mathbf{x} =	 \int_{-\infty}^{\infty} \left( \sum_{i=1}^{d}   \frac{\partial^2 f(\mathbf{x})}{\partial x_i^2} \right)^2
%\end{align}   
%%Further simplification is obtained taking $\mathbf{H}_0 = \mathbf{I}_d$. This brings AMISE to be
%%\begin{align}
%%\label{eqndamise}
%%\mbox{AMISE}({\hat{f}(\mathbf{x})}) &= \frac{h^4}{4}(\mu_2(\mathcal{K}))^2\int ( \mathbf{k}(2,d)'\mathbf{k}(2,d) ) d\mathbf{z} +  \frac{1}{Nh^d}R(\mathcal{K}) \\
%Taking derivative of $\mbox{AMISE}({\hat{f}(\mathbf{x})})$ in Equation \eqref{eqndamise} and comparing it to zero gives the optimal bandwidth parameter, minimizing the AMISE, as under: 
%\begin{align}
%\label{eqnddhamise}
%\Rightarrow h_{AMISE} &= \left( \frac{ R(\mathcal{K}) }{ \mu_2^2(\mathcal{K})R(f''(\mathbf{x}))N}\right)^{\frac{1}{4+d}}
%\end{align}
Taking partial derivative of $\mbox{AMISE}({\hat{f}(\mathbf{x})})$, in Equation \eqref{eqnddamise}, with respect $h_i$ and comparing it to zero gives the optimal bandwidth parameter $h_i$. The AMISE required is given as under: 
\begin{align}
\label{eqnddhamise}
\frac{\partial AMISE({\hat{f(\mathbf{x})}}) }{\partial h_i} &= \frac{\mu_2^2(\mathcal{K}) }{4} \left\{ h_i^3 [\mathbf{R}(f''(\mathbf{x})]_i + 2h_i\left( \sum_{j=1, j \neq i}^{d}\mathbf{h}_j^{2} \mbox{ o } \left( \left[\mathbf{R}(f''(\mathbf{x}))\right]_{r1} + \left[\mathbf{R}(f''(\mathbf{x}))\right]_{r2} \right) \right) \right\} \nonumber \\
& -  \frac{1}{ N h_i \left( \prod_{j=1}^{d} h_j \right) }R(\mathcal{K}) = 0 ; \nonumber \\
%r1 = i + \mathbf{q}_i(d^{(k-1)}+1), \mathbf{q} = (d-1):-1:1 \nonumber \\
\Rightarrow (h_{AMISE})_i &= \left( \frac{ R(\mathcal{K}) }{ \mu_2^2(\mathcal{K}) [ \mathbf{R}(f''(\mathbf{x}) ]_i \left( \prod_{j=1}^{d} h_j \right) N}\right)^{\frac{1}{4}} \nonumber \\
\Rightarrow (h_{AMISE})_i &= \left( \frac{ R(\mathcal{K}) }{ \mu_2^2(\mathcal{K}) [ \mathbf{R}(f''(\mathbf{x}) ]_i N} \right)^{\frac{1}{d+4}} \mbox{ }( \because \mbox{ Approximating det}(\mathbf{H}) = \prod_{j=1}^{d} h_j = h_i^d)
\end{align}

\hspace{0.2 in}		To simplify further the bandwidth estimation, let us assume $\mathbf{H} \in \mathcal{S}$, where $\mathcal{S} \subseteq \mathcal{D}$ with constant diagonal. Accordingly, let $\mathbf{H} = h\mathbf{I}_d$ i.e. same bandwidth in all directions. With this condition the AMISE can be given as:
\begin{align}
\label{eqndamise}
\mbox{AMISE}({\hat{f}(\mathbf{x})}) &= \frac{h^4}{4}\mu_2^2(\mathcal{K}) \left( \mathbf{I}_d^{\otimes 4}\bs{\delta}_2^{\otimes 2} \right)' \mathbf{R}(f''(\mathbf{x})) +  \frac{1}{Nh^d}R(\mathcal{K})  \nonumber \\
&= \frac{h^4}{4}\mu_2^2(\mathcal{K}) R(f''(\mathbf{x}))  +  \frac{1}{Nh^d}R(\mathcal{K}) 
\end{align}
Taking derivative of $\mbox{AMISE}({\hat{f}(\mathbf{x})})$ in Equation \eqref{eqndamise} with respect to h and comparing it to zero gives the optimal bandwidth parameter minimizing the AMISE. It is: 
%\begin{align}
%\label{eqndhamise}
%\Rightarrow h_{AMISE} &= \left( \frac{ R(\mathcal{K}) }{ \mu_2^2(\mathcal{K})R(f''(\mathbf{x}))N}\right)^{\frac{1}{4+d}}
%\end{align}
%%An optimal bandwidth parameter minimizes the total $\mbox{AMISE}({\hat{f}(\mathbf{x})})$. So,
\begin{align}
\frac{d}{dh}AMISE({\hat{f(x)}}) &= h^3(\mu_2(K))^2R(f'') -  \frac{1}{Nh^{(d+1)}}R(K) = 0  \nonumber \\
\label{eqndhamise}
\Rightarrow h_{AMISE} &= \left( \frac{R(k)}{\mu_2^2(K)R(f'')N}\right)^{\frac{1}{d + 4}}
\end{align}
Thus, the optimal bandwidth parameter depends upon some of the kernel parameters, number of samples and the second order  derivative of the actual PDF being estimated. 
\subsection{\textit{Rule-of-Thumb} for multivariate KDE} %\hspace{0.2 in} 
Let us apply the \textit{Rule-of-Thumb} for bandwidth estimation i.e. estimate the bandwidth assuming the unknown PDF as a multivariate Gaussian. For a multiplicative Gaussian kernel, $\mu_2(\mathcal{K}) = 1$,  $R(\mathcal{K}) = 2^{-d}\pi^{-d/2}$ and 
$$R(f''(\mathbf{x})) = \frac{(d+2)\left| \mathbf{C}_{\mathbf{x}}\right|^{-\frac{d+4}{2}}}{2^{(d+2)}\pi^{d/2}}.$$ So, the bandwidth parameter is obtained as 
\begin{align}
h_{ROT} = \left( \frac{4}{(2+d)N}\right)^{\frac{1}{4+d}}\sigma
\end{align}
where, $\sigma$ is the standard deviation, assumed same in all directions. 
%The same rule has been derived for different $h_j, \forall j$ in each direction, depending upon the variance in that direction.
\section{Extended Rule-of-Thumb for multivariate PDF}
\label{ndExROT}
The ExROT for bandwidth selection in multivariate KDE requires multivariate PDF approximation using infinite series. 
To estimate $R(k(2,d))$ let us first obtain $\mathbf{k(2,d)}$. 
To obtain the ExROT expressions in vector notation, let us use vector derivative operator with Kronecker product  applied twice to the Equation \eqref{GGC} for GGC Series.  
%For example, this article uses unknown PDF expressed in terms of cumulants based GCA Series. Using Equation \eqref{GCA} for GCA Series and Equation \eqref{eqndHermite} for kth order Hermite polynomial $\mathbf{H}_k$\footnote{The symbol $\mathbf{H}_k$ for kth order  Hermite polynomial need not be confused with the notation $\mathbf{H}$ for bandwidth matrix.}, $R(f''(\mathbf{x}))$ is derived as under:   
\begin{align}
\label{der2GGC}
\mathbf{k}(2,d) =  \mathbf{D}_{\mathbf{x}}^{\otimes 2}f(\mathbf{x}) = \sum_{k=0}^{\infty} (-1)^k\frac{\left( \bs{\alpha}(k,d)\otimes \mathbf{I}_{d^2}\right)'}{k!}\psi^{(k+2)}(\mathbf{x})
\end{align}
Taking Gaussian PDF as a reference PDF; i.e. $\psi(\mathbf{x})= G(\mathbf{x})$; in Equation \eqref{der2GGC} or directly taking twice differentiation of Equation \eqref{GCA} for GCA Series and Equation \eqref{eqndHermite} for kth order vector Hermite polynomial $\mathbf{H}_k$\footnote{The symbol $\mathbf{H}_k$ for kth order  Hermite polynomial need not be confused with the notation $\mathbf{H}$ for bandwidth matrix.}, $R(f''(\mathbf{x}))$ in Equation \eqref{eqndrf2} is derived as under:   
\begin{align}
%\label{eqndkf2}
\mathbf{k}(2,d) & =  \mathbf{D}_{\mathbf{x}}^{\otimes 2}f(\mathbf{x}) = \sum_{k=0}^{\infty} (-1)^k\frac{\left( \mathbf{c}(k,d)\otimes \mathbf{I}_{d^2}\right)'}{k!}G^{(k+2)}(\mathbf{x}) \\
& = \sum_{k=0}^{\infty} (-1)^k\frac{\left( \mathbf{c}(k,d)\otimes \mathbf{I}_{d^2}\right)'}{k!} \left( \mathbf{C}_\mathbf{x}^{-1}\right)^{\otimes (k+2)}\mathbf{H}_{k+2}(\mathbf{x};\mathbf{C}_{\mathbf{x}}) \\
%& \approx  \mathbf{D}_{\mathbf{x}}^{\otimes 2}f(\mathbf{x}) 
& \approx   (2\pi)^{-d/2}\left|\mathbf{C}_\mathbf{x} \right|^{-1/2}\mbox{exp}\left( -\frac{1}{2}\left( \mbox{vec} \mathbf{C}_\mathbf{x}^{-1} \right)'\mathbf{x}\otimes \mathbf{x}\right) %\nonumber \\
 \left[ \frac{(1\otimes \mathbf{I}_{d^2})'}{1}\left(\mathbf{C}_\mathbf{x}^{-1}\right)^{\otimes 2}\mathbf{H}_2(\mathbf{x}) \right. \nonumber \\
 &\mbox{ }- \left. \frac{ \left( \mathbf{c}(3,d)\otimes \mathbf{I}_{d^2}\right) '}{3!}\left(\mathbf{C}_\mathbf{x}^{-1}\right)^{\otimes 5}\mathbf{H}_5(\mathbf{x}) + \frac{\left( \mathbf{c}(4,d)\otimes \mathbf{I}_{d^2}\right) '}{4!}\left(\mathbf{C}_\mathbf{x}^{-1}\right)^{\otimes 6}\mathbf{H}_6(\mathbf{x}) \right] \nonumber \\
\label{eqndrf2ex}
R(f''(\mathbf{x})) & \approx  \sum_{i=1}^{d^4} \left[ \int_{-\infty}^{\infty}  G^2(\mathbf{x})\left\{ \left(  
 \frac{(1\otimes \mathbf{I}_{d^2})'}{1}\left(\mathbf{C}_\mathbf{x}^{-1}\right)^{\otimes 2}\mathbf{H}_2(\mathbf{x}) 
-  \frac{ \left( \mathbf{c}(3,d)\otimes \mathbf{I}_{d^2}\right) '}{3!}\left(\mathbf{C}_\mathbf{x}^{-1}\right)^{\otimes 5}\mathbf{H}_5(\mathbf{x}) 
\right. \right. \right. \nonumber \\
& \left. \left. \left. + \frac{\left( \mathbf{c}(4,d)\otimes \mathbf{I}_{d^2}\right) '}{4!}\left(\mathbf{C}_\mathbf{x}^{-1}\right)^{\otimes 6}\mathbf{H}_6(\mathbf{x}) \right) \mbox{ o } \bs{\delta}_2 \right\}^{\otimes 2}d\mathbf{x} \right]_{i}
\end{align}
The following simplification using vector algebra helps solving above Equation \eqref{eqndrf2ex}. 
\begin{align*}
& \mathbf{ab} \otimes \mathbf{cd}  = (\mathbf{a}\otimes \mathbf{c})(\mathbf{b}\otimes \mathbf{d}) \mbox{, Also } \mathbf{a}'\otimes \mathbf{b}' = (\mathbf{a}\otimes \mathbf{b})' \\
&  \left( \left( \mathbf{c}(k,d)\otimes \mathbf{I}_{d^2}\right) '\left(\mathbf{C}_\mathbf{x}^{-1}\right)^{\otimes k+2}\mathbf{H}_{k+2}(\mathbf{x})  \right)^{\otimes 2} = 
\left( \left( \mathbf{c}(k,d)\otimes \mathbf{I}_{d^2}\right)^{\otimes 2}\right)' \left( \left(\mathbf{C}_\mathbf{x}^{-1}\right)^{\otimes k+2}\mathbf{H}_{k+2}(\mathbf{x}) \right)^{\otimes 2} \\
& \left( \left( 1\otimes \mathbf{I}_{d^2}\right)'\left(\mathbf{C}_\mathbf{x}^{-1}\right)^{\otimes 2}\mathbf{H}_2(\mathbf{x}) \right)\otimes  \left( \left( \mathbf{c}(4,d)\otimes \mathbf{I}_{d^2}\right) '\left(\mathbf{C}_\mathbf{x}^{-1}\right)^{\otimes 6}\mathbf{H}_6(\mathbf{x}) \right) \\
& \hfill{		}= \left( \left( 1\otimes \mathbf{I}_{d^2}\right) \otimes \left( \mathbf{c}(4,d)\otimes \mathbf{I}_{d^2}\right) \right)' \left(\mathbf{C}_\mathbf{x}^{-1}\right)^{\otimes 8}\left(\mathbf{H}_2(\mathbf{x}) \otimes \mathbf{H}_6(\mathbf{x}) \right) \\
%\mbox{ }(\because \mathbf{ab} \otimes \mathbf{cd} = (\mathbf{a}\otimes \mathbf{c})(\mathbf{b}\otimes \mathbf{d})) \\
& \left[ \left( \left( 1\otimes \mathbf{I}_{d^2}\right)'\left(\mathbf{C}_\mathbf{x}^{-1}\right)^{\otimes 2}\mathbf{H}_2(\mathbf{x}) \right) \mbox{ o } \bs{\delta} \right]^{\otimes 2} =  \left[ \mathbf{H}_2(\mathbf{x}) \mbox{ o } \bs{\delta}_2 \right]^{\otimes 2}  = [ \left(\mathbf{C}_\mathbf{x}^{-1}\right)^{\otimes 2} \mathbf{H}_2(\mathbf{x})]^{\otimes 2} \mbox{ o } \bs{\delta}_2^{\otimes 2} %\\
%& \left[ \left( \left( \mathbf{c}(k,d)\otimes \mathbf{I}_{d^2}\right)'
%%\left(\mathbf{C}_\mathbf{x}^{-1}\right)^{\otimes k+2}
%\mathbf{H}_{k+2}(\mathbf{x})  \right) \mbox{ o } \bs{\delta}_2 \right]^{\otimes 2} = [\mathbf{p}_{d^2 \times 1}]^{\otimes 2} \mbox{ o }\bs{\delta}_2^{\otimes 2} = \mathbf{c}(k,d)^{\otimes 2}\mbox{ o } [ (\mathbf{H}_{k+2})_{i+\mathbf{q}d^2}]^{\otimes 2} \mbox{ o }\bs{\delta}_2^{\otimes 2} \\
%&\mbox{where, } \mathbf{p}_i = [\mathbf{c}(k,d)] \mbox{ o } [ (\mathbf{H}_{k+2})_{i+\mathbf{q}d^2}] \mbox{ ; } q=0:d^k-1
%%= \mbox{Mat}(\mathbf{c}(3,d))_{ii} \mbox{ o } \mox{Mat}(\mathbf{H}_{k+2})_{ii}, i=1:d^3 \mbox{ o } \bs{delta}
%%(\mathbf{a}'\mathbf{b})^2 =  (\mathbf{a}'\mathbf{b}^2 \mathbf{a})
\end{align*}
%Application of the above simplifications on Equation \eqref{eqndrf2ex} has a significant interpretation. The $R(f''(\mathbf{x}))$ and correspondingly the ExROT multivariate bandwidth parameter do not depend upon the cross-cumulants and cross-moments. This is due to the Hadamard product with the correlation matrix of the selected  product of Gaussian kernel.  

\hspace{0.2 in } The integration in equation \eqref{eqndrf2ex} is obtained using the following rules: % of integration of Gaussian.
\begin{align*}
%I = \int_{-\infty}^{\infty} \exp \left(-\mathbf{x}'\mathbf{C}_{\mathbf{x}}^{-1}\mathbf{x} \right)d\mathbf{x} &= \int_{-\infty}^{\infty} \exp \left(-\mbox{vec}\left( \mathbf{C}_{\mathbf{x}}^{-1}\right)' \mathbf{x} \otimes \mathbf{x} \right)d\mathbf{x} = \left(\frac{\pi}{\left|\mathbf{C}_{\mathbf{x}}\right|}\right)^{d/2} \\
%\mathbf{D}_{\mbox{vec} \mathbf{C}_{-1}}\int_{-\infty}^{\infty} \mathbf{x}^{2n} \exp \left(-\mathbf{x}'\mathbf{C}_{\mathbf{x}}^{-1}\mathbf{x} \right)d\mathbf{x}  &=
& I = \int_{-\infty}^{\infty} \exp \left(-\mathbf{x}'\mathbf{A}\mathbf{x} \right)d\mathbf{x}  = \int_{-\infty}^{\infty} \exp \left( -\mbox{vec} \left( \mathbf{A} \right)' \mathbf{x} \otimes \mathbf{x} \right)d\mathbf{x} 
= \left( \frac{\pi}{\left|\mathbf{A}\right|} \right)^{d/2} \\
& \Rightarrow \mathbf{D}_{\mbox{vec}(\mathbf{A})}^{\otimes } I  = \mathbf{D}_{\mbox{vec}(\mathbf{A})}^{\otimes } \left( \frac{\pi}{\left|\mathbf{A}\right|} \right)^{d/2} \\
& \Rightarrow  \int_{-\infty}^{\infty} \mathbf{x}^{\otimes 2} \exp \left(-\mbox{vec}\left( \mathbf{A} \right)' \mathbf{x} \otimes \mathbf{x} \right) d\mathbf{x}  = \frac{d}{2} \left(\frac{\pi}{|\mathbf{A}|}\right)^{d/2} \left( \mbox{vec}\mathbf{A}\right)^{\otimes (-1)}  \\
%\frac{d^n I}{d A^n} 
& \mbox{Also, } \mathbf{D}_{\mbox{vec}(\mathbf{A})}^{\otimes n} I = \mathbf{D}_{\mbox{vec}(\mathbf{A})}^{\otimes n} \left( \frac{\pi}{\left|\mathbf{A}\right|} \right)^{d/2} \\
& \Rightarrow  \int_{-\infty}^{\infty} \mathbf{x}^{\otimes 2n} \exp \left(-\mbox{vec} \left( \mathbf{A} \right)' \mathbf{x} \otimes \mathbf{x}  \right)d\mathbf{x} = \frac{(d+2(n-1))!!}{2^n} \left(\frac{\pi}{|\mathbf{A}|}\right)^{d/2} \left( \mbox{vec}\mathbf{A}\right)^{\otimes (-n)} 
\end{align*}
In general, 
\begin{align}
%\int_{-\infty}^{\infty} \mathbf{x}^{\otimes n} \exp \left(-\mbox{vec}\left( \mathbf{C}_{\mathbf{x}}^{-1} \right)' \mathbf{x} \otimes \mathbf{x} \right) d\mathbf{x}
%& =
%\begin{cases}
%\frac{(n-1)!!}{2^{n/2}} (\pi)^{d/2} \left|\mathbf{C}_{\mathbf{x}}^{-1}\right|^{-\frac{d+n}{2}} \bs{\delta}  &\mbox{ for $n$ even } \nonumber \\
%\mathbf{0} & \mbox{ for $n$ odd} % , as the function becomes odd
%\end{cases} \nonumber \\
%\mbox{where, }\bs{\delta}_i  = & 
%\begin{cases} 1 & \mbox{ ; for } i  = 1+ \mathbf{q}(d^{2n-1}+1) \mbox{ and } \mathbf{q} = 0:d-1 \nonumber \\
%0  & \mbox{ otherwise}
%\end{cases} \nonumber \\
\label{ndintGrule}
%\sum_{i=1}^{d^n}
\int_{-\infty}^{\infty} \mathbf{x}^{\otimes n} \exp \left(-\mbox{vec}\left( \mathbf{C}_{\mathbf{x}}^{-1} \right)' \mathbf{x} \otimes \mathbf{x} \right) d\mathbf{x}
& =
\begin{cases}
\frac{(d+n-2))!!}{2^{n/2}} \left( \frac{\pi}{|\mathbf{C}_{\mathbf{x}}^{-1}|}\right)^{d/2} \left( \mathbf{c}(2,d) \right)^{\otimes n/2}   &\mbox{ for $n$ even } \\
0 & \mbox{ for $n$ odd} % , as the function becomes odd
\end{cases}
\end{align}
where, $(d+n)!!= (d+n)(d+n-2)(d+n-4)\ldots d (\mbox{ or } d-1) $. The integration rule in Equation \ref{ndintGrule}  depend upon the second order vector cumulant and the length of the random vector $d$. 
%Further simplification can be obtained by assuming whiten random vector i.e. $\mathbf{C}_{\mathbf{x}}=\mathbf{I}_d$.  
%Further, the Equation \eqref{eqndrf2} also depends upon the cumulant vector. To simplify the calculations let us assume independent and identically distributed (\textit{i.i.d.}) random vector. The assumption helps in two ways. First, the independence implies uncorrelatedness i.e. $\mathbf{C}_{\mathbf{x}}=\mathbf{I}_d$. 
%%The k-order cumulant matrix $(\mathbf{C}(k,d))$ will be diagoanl with same values at the kernel. So, $\mathbf{C}(k,d) =  c_k \mathbf{I}_{d}$$\mathbf{c}(k,d) =  \mbox{vec}\mathbf{C}(k,d)$
%Second, the k-order cumculant vector will be given as,  
%\begin{align} 
%\mathbf{c}(k,d) = c_k \bs{\delta}_{d^k \times 1} \mbox{,  and }(\bs{\delta})_i =  \begin{cases} 1 \mbox{ for } i = 1 + rd, r = 0:(d-1)\\
%0 \mbox{ otherwise}
%\end{cases}
%\end{align}
%for $\forall i = 1 + rd, r = 0:(d-1)$ and all other enries will be zero. 
%\hspace{0.2 in} 
%With whitening assumption, t
The necessary integrals are obtained as under. % and also $d=2$
\begin{eqnarray}
\mathbf{t}_1 &=& \int_{-\infty}^{\infty}\exp\left( \mbox{vec} (\mathbf{C}_{\mathbf{x}}^{-1})' \mathbf{x} \otimes \mathbf{x} \right)\left[ \mathbf{H}_2(\mathbf{x}) \right]^{\otimes 2}d \mathbf{x} = \left[ \frac{d^2-2d+4}{4}\right]\left( \frac{\pi}{|\mathbf{C}_{\mathbf{x}}^{-1}|}\right)^{d/2} \mathbf{c}(2,d)^{\otimes 2}\nonumber \\
\mathbf{t}_2 &=&  \int_{-\infty}^{\infty}\exp\left( \mbox{vec} (\mathbf{C}_{\mathbf{x}}^{-1})' \mathbf{x} \otimes \mathbf{x} \right)\left[ \mathbf{H}_5(\mathbf{x}) \right]^{\otimes 2}dx \nonumber \\
&=& \left[ \frac{(d+8)!!}{2^5} + 130\frac{(d+4)!!}{2^3} + 225\frac{d}{2} - 20\frac{(d+6)!!}{2^4} - 300\frac{(d+2)!!}{2^2}\right] \left( \frac{\pi}{|\mathbf{C}_{\mathbf{x}}^{-1}|}\right)^{d/2} \mathbf{c}(2,d)^{\otimes 5} \nonumber \\
\mathbf{t}_3 &=& \int_{-\infty}^{\infty} \exp\left( \mbox{vec} (\mathbf{C}_{\mathbf{x}}^{-1})' \mathbf{x} \otimes \mathbf{x} \right) \left[ \mathbf{H}_6(\mathbf{x}) \right]{\otimes 2}dx  = \left[ \frac{(d+10)!!}{2^6} + 225\frac{(d+6)!!}{2^4} \right. \nonumber \\
 &+& \left. 2025\frac{(d+2)!!}{2^2} + 225 - 30\frac{(d+8)!!}{2^5} - 1380\frac{(d+4)!!}{2^3} - 1350\frac{d}{2}\right] \left( \frac{\pi}{|\mathbf{C}_{\mathbf{x}}^{-1}|}\right)^{d/2} \mathbf{c}(2,d)^{\otimes 6} \nonumber \\
\mathbf{t}_4 &=& \int_{-\infty}^{\infty}\exp\left( \mbox{vec} (\mathbf{C}_{\mathbf{x}}^{-1})' \mathbf{x} \otimes \mathbf{x} \right)\left[ \mathbf{H}_2(\mathbf{x})\otimes \mathbf{H}_5(\mathbf{x}) \right]dx = 0 \nonumber \\
\mathbf{t}_5 &=& \int_{-\infty}^{\infty}\exp\left( \mbox{vec} (\mathbf{C}_{\mathbf{x}}^{-1})' \mathbf{x} \otimes \mathbf{x} \right)\left[ \mathbf{H}_5(\mathbf{x})\otimes \mathbf{H}_6(\mathbf{x}) \right]dx = 0 \nonumber \\
\mathbf{t}_6 &=& \int_{-\infty}^{\infty}\exp\left( \mbox{vec} (\mathbf{C}_{\mathbf{x}}^{-1})' \mathbf{x} \otimes \mathbf{x} \right)\left[ \mathbf{H}_2(\mathbf{x})\otimes \mathbf{H}_6(\mathbf{x}) \right]dx \nonumber \\
&=& \left[ \frac{(d+6)!!}{2^4} - 16\frac{(d+4)!!}{2^3} + 60\frac{(d+2)!!}{2^2} - 60\frac{d}{2} +15 \right] \left( \frac{\pi}{|\mathbf{C}_{\mathbf{x}}^{-1}|}\right)^{d/2} \mathbf{c}(2,d)^{\otimes 4} \nonumber 
\end{eqnarray}
%Applying above simplifications to the Equation \eqref{eqndrf2} can be re-written as,
%\begin{align}
%\label{eqndrf22}
%R(f''(\mathbf{x})) & \approx \left[  T_1 + \sum_{i=1}^{d^2} \left( \mathbf{c}(3,d) \mbox{ o } \bs{\delta}  \right)^2  \frac{T_2}{d}   +  \sum_{i=1}^{d^2} \left( \mathbf{c}(4,d) \mbox{ o } \bs{\delta}  \right)^2  \frac{T_2}{d} 
%%\left|\mathbf{C}_\mathbf{x} \right|^{-3}\frac{\left( \mathbf{c}(4,d)\otimes \mathbf{I}_{d^2}\right) '}{4!}\mathbf{H}_6(\mathbf{x})\right) \mbox{ o } \bs{\delta} \right\}^{2}d\mathbf{x} \right]_{ij}
%\end{align}
Using above integrands,  the Equation \eqref{eqndrf2ex} for $R(f''(\mathbf{x}))$ can be rewritten as under:
\begin{align}
\mathbf{R}(f''(\mathbf{x})) & \approx  (2\pi)^{-d}\left|\mathbf{C}_\mathbf{x} \right|^{-1} \left\{ 
 \left( 1\otimes \mathbf{I}_{d^2}\right)^{\otimes 2'} \left(\mathbf{C}_\mathbf{x}^{-1} \right)^{\otimes 4} \mathbf{t}_1  
 +  \left( \frac{ \mathbf{c}(3,d)\otimes \mathbf{I}_{d^2} }{3!}\right)^{\otimes 2'}\left(\mathbf{C}_\mathbf{x}^{-1} \right)^{\otimes 10} \mathbf{t}_2  \right. \nonumber \\
& \left. \mbox{ }+  \left( \frac{\left( \mathbf{c}(4,d)\otimes \mathbf{I}_{d^2}\right) }{4!}\right)^{\otimes 2'}\left(\mathbf{C}_\mathbf{x}^{-1} \right)^{\otimes 12} \mathbf{t}_3  \right. \nonumber \\
& \left. + \mbox{ } \frac{\left( 1\otimes \mathbf{I}_{d^2}\right)\otimes \left( \mathbf{c}(4,d)\otimes \mathbf{I}_{d^2}\right)' }{4!}\left(\mathbf{C}_\mathbf{x}^{-1} \right)^{\otimes 8}  \mathbf{t}_6  \right\}\mbox{ o } \bs{\delta}_2^{\otimes 2} \nonumber\\
\label{eqndrf2sol}
\Rightarrow \mathbf{R}(f''(\mathbf{x})) & \approx %& =
 2^{-d}\pi^{-d/2}\left|\mathbf{C}_\mathbf{x}^{-1} \right|^{-\frac{d+2}{2}}
\left[ 
\frac{d^2-2d+4}{4} \left( \mbox{vec} \mathbf{C}_{\mathbf{x}}^{-1}\right)^{\otimes 2}	\right. \nonumber \\
& \left. \mbox{ }+ \left( \frac{(d+8)!!}{2^5} + 130\frac{(d+4)!!}{2^3} + 225\frac{d}{2} - 20\frac{(d+6)!!}{2^4} - 300\frac{(d+2)!!}{2^2} \right)  \right. \nonumber \\
& \left.  \mbox{ } \left( \frac{ \mathbf{c}(3,d)\otimes \mathbf{I}_{d^2} }{3!}\right)^{\otimes 2'}\left(\mbox{vec }\mathbf{C}_\mathbf{x}^{-1} \right)^{\otimes 5}
+ \mbox{ }\left( \frac{(d+10)!!}{2^6} + 315\frac{(d+6)!!}{2^4} + 2475\frac{(d+2)!!}{2^2}  \right.\right. \nonumber \\
& \left. \left. \mbox{ }+ 225 - 30\frac{(d+8)!!}{2^5} - 1380\frac{(d+4)!!}{2^3} - 1350\frac{d}{2}\right)
 \left( \frac{\left( \mathbf{c}(4,d)\otimes \mathbf{I}_{d^2}\right) }{4!}\right)^{\otimes 2'}\left(\mbox{vec } \mathbf{C}_\mathbf{x}^{-1} \right)^{\otimes 6}    \right. \nonumber \\
& \left. \mbox{ }
  + 2\left( \frac{(d+6)!!}{2^4} - 16\frac{(d+4)!!}{2^3} + 60\frac{(d+2)!!}{2^2} - 60\frac{d}{2} +15 \right) \right. \nonumber \\
& \left.  \mbox{ } \frac{\left( \mathbf{I}_{d^2}\otimes  \mathbf{c}(4,d)\otimes \mathbf{I}_{d^2}\right)' }{4!}\left(\mbox{vec } \mathbf{C}_\mathbf{x}^{-1} \right)^{\otimes 4}  \right] \mbox{ o }\bs{\delta}_2^{\otimes 2}
\end{align}

\hspace{0.2 in} Using  $\mu_2(\mathcal{K})=1$ and $R(\mathcal{K})=2^{-d}\pi^{-d/2}$ for d dimensional multivariate Gaussian kernel and above Equation \eqref{eqndrf2sol} for $[\mathbf{R}(f''(\mathbf{x}))]_i$ the Equation \eqref{eqnddhamise}
%, the Gram-Charlier A Series based ExROT for bandwidth selection in multivariate KDE 
is obtained. 
The total roughness $R(f''(\mathbf{x})) = \mathbf{R}(f''(\mathbf{x}))'\bs{\delta}_2^{\otimes 2}$. For better visulization, there is assumed whiten random vector with $\mathbf{C}_{\mathbf{x}} = \mathbf{I}_d$ or $\mbox{vec}\mathbf{C}_{\mathbf{x}} = \bs{\delta}_2$. 
% Also, let symbolically, $\left( \mbox{devec}_{(d^4,d^{2k})} \left( \mbox{vec}(\mathbf{C}_{\mathbf{x}}^{-1})^{\otimes k+2} \right)\mathbf{c}(k,d)^{\otimes 2} \right)' \bs{\delta}_2^{\otimes 2} = p_k$ and $\left( \mbox{devec}_{(d^4,d^{2k})} \left( \mbox{vec}(\mathbf{C}_{\mathbf{x}}^{-1})^{\otimes k+2} \right)\mathbf{c}(k,d)^{\otimes 2} \right)' \bs{\delta}_2^{\otimes 2} = p_k$. 
The further simplification is achieved as under:
\begin{align*}
&\mbox{As, } \mathbf{ABd} = (d' \otimes \mathbf{A}) \mbox{vec} \mathbf{B} \\
&\left( \mathbf{c}(k,d)\otimes \mathbf{I}_{d^2}\right)^{\otimes 2'}
\left(\mbox{ vec }\mathbf{C}_\mathbf{x}^{-1}\right)^{\otimes k+2} = \mbox{devec}_{(d^4,d^{2k})} \left( \mbox{vec}(\mathbf{C}_{\mathbf{x}}^{-1})^{\otimes k+2} \right) %\bs{\delta}_2^{\otimes k+2} 
 \mathbf{c}(k,d)^{\otimes 2} \\ %= p(k,d) (symbolically) \\
& \Rightarrow\left(  \left( \mathbf{c}(k,d)\otimes \mathbf{I}_{d^2}\right)^{\otimes 2'}
\left(\mbox{ vec }\mathbf{C}_\mathbf{x}^{-1}\right)^{\otimes k+2} \right) \mbox{ o } \bs{\delta}_2 = \mathbf{c}(k,d)^{\otimes 2'}\bs{\delta}_2^{\otimes k} \mbox{ }(\because \mbox{vec}\mathbf{C}_{\mathbf{x}} = \bs{\delta}_2)\\
& \mbox{Also, } \left( \left( \mathbf{I}_{d^2}\otimes  \mathbf{c}(4,d)\otimes \mathbf{I}_{d^2}\right)' \left(\mbox{vec } \mathbf{C}_\mathbf{x}^{-1} \right)^{\otimes 4}  \right)'\bs{\delta}_2^{\otimes 2} = \mathbf{c}(4,d)'\bs{\delta}_2^{\otimes 2} 
%\mathbf{H}_{k+2}(\mathbf{x})  \right) \mbox{ o } \bs{\delta}_2 \right]^{\otimes 2} = [\mathbf{p}_{d^2 \times
%Overall, the  and multivariate Gaussian kernel. 
\end{align*}
where, devec is an operation converting a vetor into matrix as the dimensions specified.   Then, the bandwidth selection rule in Equation \eqref{eqndhamise} as under.
%Overall, the ExROT for multivariate KDE of a whiten random vector with Gaussian product kernel is derived as under: 
\begin{align}
h_{GC} &= (CN)^{-\frac{1}{d+4}}  \\
\mbox{where,  }
C &= \left[ 
\frac{d^2-2d+4}{4} \right. \nonumber \\
& \left. + \frac{1}{d}\left( \frac{(d+8)!!}{2^5} + 130\frac{(d+4)!!}{2^3} + 225\frac{d}{2} - 20\frac{(d+6)!!}{2^4} - 300\frac{(d+2)!!}{2^2} \right)  \left( \frac{\mathbf{c}(3,d)^{\otimes 2'}\bs{\delta}_2^{\otimes 3} }{3!^2}\right) \right. \nonumber \\
& \left. + \frac{1}{d}\left( \frac{(d+10)!!}{2^6} + 315\frac{(d+6)!!}{2^4} + 2475\frac{(d+2)!!}{2^2} + 225 - \right.\right. \nonumber \\
& \left. \left. 30\frac{(d+8)!!}{2^5} - 1380\frac{(d+4)!!}{2^3} - 1350\frac{d}{2}\right)
 \left( \frac{\mathbf{c}(4,d)^{\otimes 2'}\bs{\delta}_2^{\otimes 4}}{4!^2} \right)   \right. \nonumber \\
& \left. 
  + 2\frac{1}{d}\left( \frac{(d+6)!!}{2^4} - 16\frac{(d+4)!!}{2^3} + 60\frac{(d+2)!!}{2^2} - 60\frac{d}{2} +15 \right) \frac{\mathbf{c}(4,d)'\bs{\delta}_2^{\otimes 2}}{4!} \right] \nonumber 
\end{align}
%where, $\mathbf{1}_{d^4}$ is the $d^4 \times 1$ dimensional vector of ones. 
%The above simpliefied bandwidth selection rule is for KDE of d-variate whiten random vector.  

\hspace{0.2 in}	Over all, the Gram-Charlier A Series based \textit{Extended Rule-of-Thumb} for bandwidth parameter $h_{GC}$ selection in multivariate KDE is obtained. The derivation assumes near Gaussian unknown PDF and Gaussian kernel.
% and whiten random vector. 
%The derived bandwidth selection rule  for dimension d is given as under. 
With $\mathbf{c}(3,d)= 0,\mathbf{c}(4,d) =0$, the Silverman's \textit{Rule-of-Thumb} is one case of the extended rule.

\hspace{0.2 in} For better visualization, the same equation for $d=2$ and whiten random vector is given as under: 
\begin{eqnarray}
R(f''(\mathbf{x})) &=& \frac{1}{4\pi} \left[ 1 + 45 \left( \frac{\mathbf{c}(3,d)^{\otimes 2'}\bs{\delta}_2^{\otimes 3}}{3!^2}\right) + 225 \left( \frac{\mathbf{c}(4,d)^{\otimes 2'}\bs{\delta}_2^{\otimes 4}}{4!}\right)  + 6 \frac{ \mathbf{c}(4,d)'\bs{\delta}_2^{\otimes 2}}{4!} \right] \nonumber \\
\label{eqndrf22d}
R(f''(\mathbf{x})) &=& \frac{1}{4\pi} \left[ 1 + 1.25 \mathbf{c}(3,d)^{\otimes 2'}\bs{\delta}_2^{\otimes 3} + 0.3906 \mathbf{c}(4,d)^{\otimes 2'}\bs{\delta}_2^{\otimes 4} + 0.25 \mathbf{c}(4,d)'\bs{\delta}_2^{\otimes 2} \right] 
\end{eqnarray}
With $R(f''(\mathbf{x}))$ from Equation \eqref{eqndrf22d}, $R(\mathcal{K})= 2^{-d}\pi^{-d/2}$ and $\mu_2(\mathcal{K})=1$ for Gaussian kernel gives the required bandwidth parameter from Equation \eqref{eqndhamise} is given as:
\begin{eqnarray}
h_{GC} &=& (CN)^{-\frac{1}{6}}  \\
\mbox{where,  }
C &=&
\begin{cases}
  1  \hfill{ }\mbox{ if both }\mathbf{c}(3,d)= \mathbf{0}, \mathbf{c}(4,d) &= \mathbf{0};  \nonumber \\ %i.e. Gaussian PDF
 1 + 0.3906 \mathbf{c}(4,d)^{\otimes 2'}\bs{\delta}_2^{\otimes 4}  + 0.25 \mathbf{c}(4,d)'\bs{\delta}_2^{\otimes 2}  \hfill{ }\mbox{ if } \mathbf{c}(3,d)&= \mathbf{0}; \nonumber\\%i.e. symmetric PDF
 %1 + 6.5625 k_3 + 2.9821 k_4 	& \text{if $\sigma=1$}, \nonumber\\
% 1 + 2.9821k_4 	& \text{if both $k_3=0$ and $\sigma=1$}, \nonumber\\
 1 + 1.25 \mathbf{c}(3,d)^{\otimes 2'}\bs{\delta}_2^{\otimes 3} + 0.3906 \mathbf{c}(4,d)^{\otimes 2'}\bs{\delta}_2^{\otimes 4}  + 0.25 \mathbf{c}(4,d)'\bs{\delta}_2^{\otimes 2}  & \mbox{otherwise} \nonumber
\end{cases}
%\begin{cases}
%1.0592\sigma \left(\left( 1 + 2.2559\frac{k_4}{\sigma^8} + 0.7292\frac{k_4}{\sigma^{4}}\right)N\right)^{-\frac{1}{5}} & \text{ if $k_3 = 0$ i.e. symmetric PDF}, \\
%1.0592 \left(\left( 1 + 6.5625 k_3 + 2.9821 k_4 \right)N\right)^{-\frac{1}{5}}	& \text{if $\sigma=1$}, \\
%1.0592 \left(\left( 1 + 2.9821k_4 \right)N\right)^{-\frac{1}{5}}	& \text{if both $k_3=0$ and $\sigma=1$}, \\
%1.0592\sigma \left(\left( 1 + 6.5625\frac{k_3}{\sigma^6} + 2.2559\frac{k_4}{\sigma^8} + 0.7292 \frac{k_4}{\sigma^{4}}\right)N\right)^{-\frac{1}{5}} & \text{otherwise}
%\end{cases}
\end{eqnarray}
%It is worth reminding that the $p(k,d)$ depends upon the second order Kronecker product of kth order vector cumulants $\mathbf{c}(k,d)$ and $q(4,d)$ depends on the $\mathbf{c}(4,d)$. 
%Empirically, the absolute value of $C$ is considered to avoid the imaginary result.
\section[ExROT for Density Derivative Estimator]{ExROT for Bandwidth Selection in  Kernel Density Derivative Estimator}
\label{ExROT4Der}
%{ExROT for Bandwidth Selection in Univariate and Multivariate Kernel Density Derivative Estimator}
In general, the $r^{th}$ derivative of univariate or multivariate density $f(\mathbf{x})$ using a $v^{th}$ order kernel is estimated as  under:
\begin{align}
\label{eqgradkde}
 \hat{f}^{(r)}(\mathbf{x}) = \frac{1}{N}\sum_{i=1}^{N}{\mathcal{K}_{v,\mathbf{H}}^{(r)}\left(\mathbf{x}-\mathbf{x}_i\right)} 
\end{align}
where, usually a product kernel is used with whitening in all directions and $\mathbf{H}= h\mathbf{I}_{d \times d}$ as defined under:
\begin{align}
\mathcal{K}_{v,\mathbf{H}}^{(r)}(\mathbf{x}-\mathbf{x}_i) = \prod_{j=1}^{d}\frac{1}{h}\mathcal{K}_v^{(r_j)}\left( \frac{x_{ji}-x_j}{h}\right)
\end{align}
The $(r_j)^{th}$ derivative corresponds to the fact that
\begin{align*}
f^{(r)}(x) = \frac{\partial^{r} f(x)}{\partial^{r_1}x_1\partial^{r_2}x_2\ldots \partial^{r_d}x_d}
\end{align*}
The AMISE for density derivative as derived in \citep{henderson2012normal} and from Equation \ref{eqndamise} is given as under:
\begin{align}
AMISE\left\{ \hat{f}^{(r)}(\mathbf{x}) \right\} &= \frac{ \mu_2(\mathcal{K}_v) }{(v!)^2} \int h^{2v} \left( \mathbf{k}(r+v,d)'\left( \mathbf{I}_d^{\otimes 2}\bs{\delta}_2\right) \right)^2 d\mathbf{x} +  \frac{1}{Nh^{d+2r}}R(\mathcal{K}^{(r)}) \\
\label{eqndgethamise}
\Rightarrow h_{AMISE} &= \left( \frac{(v!)^2}{2v} \frac{ R(\mathcal{K}^{(r)}) }{ \mu_2^2(K) R( f^{(r+v)}(\mathbf{x} ) N } 	 \right)^{\frac{1}{2v +2r +d}}
\end{align}
%where, $R(\mathcal{K}^{(r)}) = h^{-(2+2r)}\prod_{j=1}^{d}R(\mathcal{K}^{r_j})$.

\hspace{0.2 in}		With this, the ExROT for gradient density can be derived and needs $R(\mathbf{k}(r+v,d))$ definition.
With Gaussian kernel i.e. $v=2$ and $r=1$ the required parameter for 1-dimension can be derived as under.
\begin{align}
R(f^{(3)}(x)) &= \int_{\infty}^{\infty}{\frac{1}{2\pi\sigma}\mbox{exp}\left( -\left(z\right)^2\right)\left[ - \frac{1}{\sigma^3}H_3(z) + \frac{k_3}{3!\sigma^6}H_6(z) - \frac{k_4}{4!\sigma^7}H_7(z) \right]^2dz}  \nonumber \\
&= \frac{1}{2\sqrt{\pi}\sigma}\left[\frac{1}{\sigma^6}\frac{15}{8} + \frac{1}{\sigma^{12}} \left(\frac{k_3}{3!}\right)^2\frac{10395}{64} + \frac{1}{\sigma^{14}} \left(\frac{k_4}{4!}\right)^2\frac{135135}{2^7}   + \frac{1}{\sigma^{9}} \left(\frac{k_4}{4!}\right)\frac{945}{32} \right]  \nonumber \\
%\end{align}
%Accordinglly, the bandwidth $h_{GC}$ for density derivative for $d=1$ can be given as under:
%\begin{eqnarray}
\Rightarrow h_{GC} &= \sigma (CN)^{-\frac{1}{7}} \\
\mbox{where,  }
C &= \frac{1.875}{\sigma^6} + 4.5117\frac{k_3^2}{\sigma^{12}} + 1.8329\frac{k_4^2}{\sigma^{14}} + 2.4609 \frac{k_4}{\sigma^{6}}
%\begin{cases}
%  1 & \text{both $k_3=K_4=0$ i.e. Gaussian PDF} \nonumber \\
% 1 + 2.2559\frac{k_4}{\sigma^8} + 0.7292\frac{k_4}{\sigma^{4}} & \text{ if $k_3 = 0$ i.e. symmetric PDF}, \nonumber\\
% 1 + 6.5625 k_3 + 2.9821 k_4 	& \text{if $\sigma=1$}, \nonumber\\
% 1 + 2.9821k_4 	& \text{if both $k_3=0$ and $\sigma=1$}, \nonumber\\
% 1 + 6.5625\frac{k_3}{\sigma^6} + 2.2559\frac{k_4}{\sigma^8} + 0.7292 \frac{k_4}{\sigma^{4}} & \text{otherwise} \nonumber
%\end{cases}
\end{align}
Similarly,  with Gaussian kernel i.e. $v=2$,  $r=1$ and whiten components the bandwidth parameter for d-dimensional density derivative  is derived as under:
\begin{align}
h_{GC} &= (CN)^{-\frac{1}{d+2+4}}  \\
\mbox{where,  }
C &= \left[ 
\left( \frac{(d+4)!!}{2^3} + 9 \frac{d}{2} - 6 \frac{(d+2)!!}{2^2}\right) \right. \nonumber \\
& \left. +\left( \frac{(d+10)!!}{2^6} + 315\frac{(d+6)!!}{2^4} + 2475\frac{(d+2)!!}{2^2} + 225 - \right.\right. \nonumber \\
& \left. \left. 30\frac{(d+8)!!}{2^5} - 1380\frac{(d+4)!!}{2^3} - 1350\frac{d}{2}\right)
 \left( \frac{\mathbf{c}(3,d)^{\otimes 2'}\bs{\delta}_2^{\otimes 3}}{3!^2} \right)   \right. \nonumber \\
& \left. + \left( \frac{(d+12)!!}{2^7}+ 651\frac{(d+8)!!}{2^5} + (105^2+4410) \frac{(d+4)!!}{2^3} + 105^2\frac{d}{2} - 4620\frac{(d+6)!!}{2^4}  \right. \right. \nonumber \\
&\left. + \left. - 2*105^2 \frac{(d+2)!!}{2^2} - 42\frac{(d+10)!!}{2^6} \right)  \left( \frac{\mathbf{c}(4,d)^{\otimes 2'}\bs{\delta}_2^{\otimes 2}}{4!^2}\right) \right. \nonumber \\
& \left. 
  + \left( \frac{(d+8)!!}{2^5} - 24\frac{(d+6)!!}{2^4} + 168\frac{(d+4)!!}{2^3} - 420\frac{(d+2)!!}{2^2} + 315\frac{d}{2} \right) \frac{\mathbf{c}(4,d)'\bs{\delta}_2^{\otimes 2}}{4!} \right] \nonumber 
\end{align}
For better visualization, the same equation  for $d=2$ and whiten random vector is given as under:
\begin{eqnarray}
R(f''(\mathbf{x})) &=& \frac{1}{4\pi} \left[ 3 + 225 \left( \frac{\mathbf{c}(3,d)^{\otimes 2'}\bs{\delta}_2^{\otimes 3}}{3!^2}\right) + \frac{201600}{2^7} \left( \frac{\mathbf{c}(4,d)^{\otimes 2'}\bs{\delta}_2^{\otimes 4}}{4!^2}\right)  + \frac{864}{32} \frac{\mathbf{c}(4,d)'\bs{\delta}_2^{\otimes 2}}{4!} \right] \nonumber \\
h_{GC} &=& (CN)^{-\frac{1}{8}}  \\
\mbox{where,  }
C &=& 3 + 6.25 \mathbf{c}(3,d)^{\otimes 2'}\bs{\delta}_2^{\otimes 3} + 2.7344 \mathbf{c}(4,d)^{\otimes 2'}\bs{\delta}_2^{\otimes 4} + 2.25 \mathbf{c}(4,d)'\bs{\delta}_2^{\otimes 2}
%\begin{cases}
%  1 & \text{both $\mathbf{c}(3,2)= 0,\mathbf{c}(4,2) =0$ } \nonumber \\ %i.e. Gaussian PDF
% 1 - 4.9219 \mathbf{c}^2(4,d)  + 0.1250 \mathbf{c}(4,d)'*\mathbf{1}_{d^4} & \text{ if $\mathbf{c}(3,2) = 0$ }, \nonumber\\%i.e. symmetric PDF
% %1 + 6.5625 k_3 + 2.9821 k_4 	& \text{if $\sigma=1$}, \nonumber\\
%% 1 + 2.9821k_4 	& \text{if both $k_3=0$ and $\sigma=1$}, \nonumber\\
% 1 + 1.25 \mathbf{c}^2(3,d) - 4.9219 \mathbf{c}^2(4,d)  + 0.1250 \mathbf{c}(4,d)'*\mathbf{1}_{d^4}  & \text{otherwise} \nonumber
%\end{cases}
\end{eqnarray} 
\section{Conclusion and Future directions}
\label{bwkdeconclusion} 
The article addresses the issue of bandwidth selection in KDE for both - univariate and multivariate. There has been proposed Gram-Charlier A Series based \textit{Extended Rule-of-Thumb} (ExROT) 
% on the PDF expansion through 
%infinite series. The concept is demonstrated using PDF expansion through . The  which approximates the unknown PDF in terms its cumulants 
on the assumption that the density being estimated is near Gaussian. The performance analysis of ExROT is  done using standard test set for univariate density estimation. The results show that in all nongaussian unimodal density estimation cases - skewed or kurtotic or with outlier - ExROT has performed better than ROT. This is achieved at  %marginally increased 
computation comparable to ROT and  too small compare to the $\epsilon$-exact solve-the-equation plug-in rule. 
%The worst performance of ExROT in  multimodal density estimation is due to wrong estimation of the  skewness and kurtosis. Still, 
The ExROT  has also outperformed ROT  in some of the skewed multimodal density estimation -  skewed bimodal, claw, Asymmetric claw. The Gram-Charlier A Series based ExROT for bandwidth selection is also obtained for multivariate KDE and multivariate density derivative estimations.
%The ExROT in multivariate  KDE requires multivariate Gram-Charlier series. To match this need, first a specific derivation for univariate generalized Gram-Charlier series is extended for multivariate using Kronecker product. 
%Overall, ExROT surely is a better option to ROT in unimodal density estimation. It is also a promising option for multimodal density estimation,  as better approximation of kutosis and skewness or may be better approximation of the multimodal densities through series expansion can improve further the performances. The next article uses the derived 
%ExROT for mainly two purposes: 1) to estimate the previously proposed BSS contrast and  2) to estimate the ITL based other quantities. In both cases the interesting query is how much the estimation is improved due to the use of ExROT then the simple ROT? Then, the question is how much precision obtained and can it be treated parameter free. 

\hspace{0.2 in} The article serves as a particular demonstration to a more generalized class of bandwidth selection rules based on PDF approximations through infinite series. 
The PDF approximation through infinite series expansion is a well established area and there exist many such approximations based on various reference PDFs. As the first results are encouraging, many such rules can be developed.  % based on class of ma. 
%pproximations be tried with Gaussian or other reference PDFs. 
%Also, it seems possible to use ExROT as a first guess for the more precise \textit{solve-the-equation  plug-in} rules.
\appendix
\section{AMISE for bandwidth parameter selection in KDE}
\label{appamise1d}
 Given N realizations of an unknown PDF $f(x)$, the kernel density estimate ${\hat{f(x)}}$ is given by %The problem is to estimate $f^{^}(x)$. 
 \begin{equation}
 \label{apeqkdech3}
 \hat{f(x)} = \frac{1}{Nh}\sum_{i=1}^{N}{K\left(\frac{x-x_i}{h}\right)} = \frac{1}{N}\sum_{i=1}^{N}{K_h\left(x-x_i\right)}
 \end{equation}
 where, $K(u)$ is the kernel function and h is the bandwidth parameter. Usually, $K(u)$ is a symmetric, positive definite and bounded function;  mostly a PDF;  satisfying the following properties:  %i.e. it satisfies the following properties:
 \begin{equation*}
 K(u) \geq 0, \mbox{  }\int_{-\infty}^{\infty}{K(u)du} = 1, \mbox{  }\int_{-\infty}^{\infty}{uK(u)Du} = 0, \mbox{  }\int_{-\infty}^{\infty}{u^2K(u)du} = \mu_2(K) < \infty 
 \end{equation*}
 The accuracy of a PDF estimation can be quantified by the available distance measures between PDFs; like; $L_1$ norm based mean integrated absolute measure, $L_2$ norm based mean integrated square error (MISE), Kullback-Libeler divergence and others. The optimal smoothing parameter (the bandwidth) $h$ is obtained by minimizing the selected distance measure. The bandwidth selection rule based on the most widely used criteria MISE or IMSE (Integrated Mean Square Error) is derived as under \citep{Silverman86,wand1994kernel}. 
 \begin{eqnarray}
 \mbox{ISE}(f(x),{\hat{f(x)}}) &=& L_2(f(x),{\hat{f(x)}}) := \int_{-\infty}^{\infty}{({\hat{f(x)}}-f(x))^2 dx} \nonumber \\
 \mbox{MISE}( f(x),{\hat{f(x)}} ) &=&  E\{ ISE( f(x),{\hat{f(x)}} )\} = E\left\{ \int_{-\infty}^{\infty}{ ( {\hat{f(x)}}-f(x) )^2 dx }\right\}  \nonumber \\
 &=&  \int_{-\infty}^{\infty}{ E\{ ({\hat{f(x)}}-f(x))^2 \} dx} = \int_{-\infty}^{\infty}{ \mbox{MSE}( f(x),{\hat{f(x)}} ) } = \mbox{IMSE}( f(x),{\hat{f(x)}} )  \nonumber \\
 &=& \int_{-\infty}^{\infty}{ ( E\{\hat{f(x)}\}-f(x) )^2 + E\{ ( {\hat{f(x)}} - E\{\hat{f(x)}\} )^2 \} dx}  \nonumber \\
\label{apeqmise} 
 &=& \int_{-\infty}^{\infty}{\mbox{Bias}^2({\hat{f(x)}})dx} + \int_{-\infty}^{\infty}{\mbox{Var}({\hat{f(x)}})dx} 
 \end{eqnarray}
%Now,
\begin{eqnarray}
\mbox{Now, }
E\{\hat{f(x)}\} &=& \frac{1}{Nh}\sum_{i=1}^{N}{E\left\{K\left(\frac{x-x_i}{h}\right)\right\} }	 \nonumber \\
%&=& \frac{1}{Nh}\sum_{i=1}^{N}{ \int_{-\infty}^{\infty}{K\left(\frac{x-s}{h}\right)f(s)ds}}	 \nonumber \\
&=& \frac{1}{h}\int_{-\infty}^{\infty}{  K\left(\frac{x-s}{h}\right) f(s)ds }	 \nonumber \\
&=&   \int_{-\infty}^{\infty}{ K(z)f(x-hz)dz } \mbox{   ( $\because$ substituting $z=\frac{x-s}{h}$ ) }  \nonumber 
%&=&   \int_{-\infty}^{\infty}{ K(z)\left( f(x) - hzf'(x) + \frac{h^2z^2}{2}f''(x) + O(h^2) \right) dz } \nonumber
\end{eqnarray}
%\mbox{   ($\because$ Expanding $f(x-hz)$ using Taylor series assuming $f(x)$ a smooth density i.e. all derivatives exist)}  \nonumber \\
Expanding $f(x-hz)$ using Taylor Series with the assumption of $f(x)$ being a smooth PDF i.e. all derivatives exist
%\begin{align}
%E\{\hat{f(x)}\} =  \int_{-\infty}^{\infty}{ K(z)\left( f(x) - hzf'(x) + \frac{h^2z^2}{2}f''(x) + O(h^2) \right) dz } \nonumber
%\end{align}
\begin{eqnarray}
E\{{\hat{f(x)}}\} &=& \int_{-\infty}^{\infty}{ K(z)\left( f(x) - hzf'(x) + \frac{h^2z^2}{2}f''(x) + O(h^2) \right) dz }\nonumber \\
&=& f(x) +\frac{h^2}{2}\mu_2(K)f''(x) + O(h^2)	 \nonumber \\
\label{apeqbias}
\Rightarrow \mbox{Bias}({\hat{f(x)}}) &\approx& \frac{h^2}{2}\mu_2(K)f''(x) 	
\end{eqnarray}
where, $\mu_2(K) = \int{z^2K^2(z)dz}$, the second order central moment of the kernel. 
\begin{eqnarray}
\mbox{Similarly, }
\mbox{Var}({\hat{f(x)}}) &=& \mbox{Var}\left( \frac{1}{Nh}\sum_{i=1}^{N}{K\left(\frac{x-x_i}{h}\right)}\right)  \nonumber \\
&=& \frac{1}{N^2h^2} \sum_{i=1}^{N}{ \mbox{Var}\left( K\left(\frac{x-x_i}{h}\right)\right)}  \nonumber \\
&=& \frac{1}{Nh^2}\left[E\left\{ K^2\left(\frac{x-x_i}{h}\right) \right\} - E^2 \left\{ K\left(\frac{x-x_i}{h}\right)\right\} \right]	 \nonumber \\ 
&=&  \frac{1}{N}\int_{-\infty}^{\infty}{ \frac{1}{h^2} \left( K^2\left(\frac{x-s}{h}\right)\right)f(s)ds } -  \frac{1}{N}\left( \frac{1}{h}\int_{-\infty}^{\infty}{K\left(\frac{x-s}{h}\right)f(s)ds }\right)^2	 \nonumber \\
%\left( f(x) + \mbox{Bias}(\hat{f(x)})\right)^2 \\
%\mbox{Var}({\hat{f(x)}}) &=& E\left[ \left({\hat{f(x)}}-E[{\hat{f(x)}}]\right)^2\right] \\
%%= E\left[({\hat{f(x)}})^2\right]- \left(E[{\hat{f(x)}}]\right)^2\\%\mbox{Var}\left( \frac{1}{Nh}\sum_{i=1}^{N}K\left(\\frac{x-x_i}{h}right)\right) 
%&=& E\left[ \left(\frac{1}{Nh}\sum_{i=1}^{N}{K\left(\frac{x-x_i}{h}\right)}\right)^2\right] - \left(E \left[  \frac{1}{Nh}\sum_{i=1}^{N}{K\left(\frac{x-x_i}{h}\right)} \right]\right)^2\\
%%E\left[\left( f(x) + \mbox{Bias}(\hat{f(x)})\right)^2 \right]\\
%&=&  \frac{1}{N}\int_{-\infty}^{\infty}{ \frac{1}{h^2} \left( K\left(\frac{x-s}{h}\right)\right)^2f(s)ds } -  \frac{1}{N}\left( \frac{1}{Nh}\sum_{i=1}^{N}{E\left[K\left(\frac{x-x_i}{h}\right)\right] }\right)^2	\\%\left( f(x) + \mbox{Bias}(\hat{f(x)})\right)^2 \\
&=& \frac{1}{Nh}\int_{-\infty}^{\infty}{  K^2(z) f(x-hz)dz }  - \frac{1}{N}\left( f(x) + \mbox{Bias}(\hat{f(x)})\right)^2 
%%\left( f(x) + \frac{h^2z^2}{2}f''(x) + O(h^2) \right)^2 
%\mbox{   ( $\because$ substituting $z=\frac{x-s}{h}$ ) }  \nonumber 
\end{eqnarray}
where, s is the mean of x and $z=\frac{x-s}{h}$. 
Now, expanding $f(x-hz)$ using Taylor Series with assumption of $f(x)$ being a smooth PDF i.e. all derivatives exist
\begin{eqnarray} 
\mbox{Var}( {\hat{f(x)} }) &=& \frac{1}{Nh}\int_{-\infty}^{\infty}{ K^2(z)\left( f(x) - hzf'(x) + \frac{h^2z^2}{2}f''(x) + O(h^2) \right) dz } \nonumber \\
&& - \frac{1}{N}\left( f(x) + \frac{h^2z^2}{2}f''(x) + O(h^2) \right)^2  \nonumber \\
\label{apeqvar}
\Rightarrow \mbox{Var}( {\hat{f(x)} }) &\approx& \frac{1}{Nh}f(x)\int{K^2(z)dz}  \mbox{    ($\because$ assuming large $N$, small $h$ ) }% and given kernel properties
%\Rightarrow \mbox{Var}( {\hat{f(x)} }) &\approx& \frac{1}{Nh}f(x)\int{K^2(z)dz}
 \end{eqnarray}
Combining equations \eqref{apeqmise}, \eqref{apeqbias} and \eqref{apeqvar}
\begin{equation}
\mbox{MISE}({\hat{f(x)}}) = \frac{h^4}{4}(\mu_2(K))^2R(f'') +  \frac{1}{Nh}R(K) + O(h^4) + O\left( \frac{h}{N} \right) \nonumber 
\end{equation}
where, %$\mu_2(K) = \int{z^2K^2(z)dz}$,  
$R(f'') = \int{(f''(x))^2dx}$ and $R(K) = \int{K^2(z)dz}$. In general, $R(g) = \int{g^2(z)dz}$ is identified as the roughness of function $g(x)$. 
An asymptotic large sample approximation AMISE is obtained, assuming $\lim_{N\rightarrow\infty}{h} = 0$ and $\lim_{N\rightarrow\infty}{Nh} = \infty$ i.e. h reduces to 0 at a rate slower than $1/N$.
\begin{equation}
\label{apeqmisef}
\mbox{AMISE}({\hat{f(x)}}) = \frac{h^4}{4}(\mu_2(K))^2R(f'') +  \frac{1}{Nh}R(K) 
\end{equation}
The Equation \eqref{apeqmisef} interprets that a small $h$ increases estimation variance, whereas, a larger $h$ increases estimation bias.
% reduces estimation bias but increases estimation variance and a larger $h$ reduces estimation variance but  increases estimation bias.
An optimal $h$ minimizes the total $\mbox{AMISE}({\hat{f(x)}})$. So,
\begin{eqnarray}
\frac{d}{dh}AMISE({\hat{f(x)}}) &=& h^3(\mu_2(K))^2R(f'') -  \frac{1}{Nh^2}R(K) = 0  \nonumber \\
\label{apeqhamise}
\Rightarrow h_{AMISE} &=& \left( \frac{R(k)}{\mu_2(K)^2R(f'')N}\right)^{\frac{1}{5}}
\end{eqnarray}
Thus, the optimal bandwidth parameter depends upon some of the kernel parameters, number of samples and the second derivative of the actual PDF being estimated. 
\section{Kronecker Product and K-Derivative Operator}
\label{KronDer}
The section briefs preliminary requisites on the Kronecker product and its repeated applications to the vector  differential operator. More details can be found in \cite{Tensor87,ndHermite02,SankhyaCumVec06}.
\begin{definition}[Kronecker Product Operator $(\otimes)$]
The Kronecker Product Operator $(\otimes)$ between matrices $\mathbf{A}$ with size $p \times q$ and $\mathbf{B}$ with size $m \times n$ is defined as:\\
$$\mathbf{A}\otimes\mathbf{B} =  \left[
\begin{array}[]{c c c c}
a_{11}\mathbf{B} & a_{12}\mathbf{B} & \cdots & a_{1q}\mathbf{B} \\
a_{21}\mathbf{B} &a_{22}\mathbf{B} &\cdots &a_{2q}\mathbf{B} \\
\vdots & \vdots & \ddots&\vdots \\
a_{p1}\mathbf{B} &a_{p2}\mathbf{B} &\cdots &a_{pq}\mathbf{B}	
\end{array}
\right]$$
\end{definition}
The resultant matrix is of dimension $pm \times qn$. 
%The $Vec$ operator converts $mp \times nq$ matrix into a $ mpnq \times 1$ column vector. 
 
\hspace{0.2 in} Let $\mathbf{A}$ is with size $p \times 1$ and $\mathbf{B}$ is with size $m \times 1$, then\footnote{The symbol ' stands for Transpose of a matrix} $\mathbf{A} \otimes \mathbf{B}'$ is a matrix with size $p \times m $.  $\mathbf{A} \otimes \mathbf{A}$ is symbolically represented as $\mathbf{A}^{\otimes 2}$ and has size $p^2 \times 1$. In general, $\mathbf{A} \otimes \mathbf{A} \otimes \ldots \otimes \mathbf{A} \mbox{ (n times)}$ is symbolically represented as $\mathbf{A}^{\otimes n}$ and has size $p^n \times 1$.  Accordingly, $\mathbf{A}^{\otimes n}B$  has size $p^nm \times 1$.  
%With this much about the Kronecker product, let the derivation start with introducing multivariate Taylor series.
\begin{definition}[Jacobian Matrix] 
Let $\bs{\lambda} = (\lambda_1, \lambda_2, \ldots, \lambda_d)'$, $\bs{\lambda} \in \mathbb{R}^d$ and  $\mathbf{f}(\bs{\lambda}) = $ \\ 
$(f_1(\lambda), f_2(\lambda), \ldots, f_m(\lambda))' \in \mathbb{R}^{m}$ be a differentiable m-component vector function. Then, Jacobian matrix $(\mathbf{J})$ of $\mathbf{f}$  $(\mathbf{J}(\mathbf{f}))$ is an $m \times d$ matrix defined as under:
\begin{align*}
\mathbf{J}(\mathbf{f}(\bs{\lambda})) = \frac{\partial \mathbf{f}}{\partial {\bs{\lambda}}} = \left[ \frac{\partial \mathbf{f}}{\partial \lambda_1}, \frac{\partial \mathbf{f}}{\partial \lambda_2},\ldots, \frac{\partial \mathbf{f}}{\partial \lambda_d} \right] = \left[
\begin{array}[]{c c c c}
\frac{\partial f_1}{\partial \lambda_1} & \frac{\partial f_1}{\partial \lambda_2}  & \cdots & \frac{\partial f_1}{\partial \lambda_d} \\
\frac{\partial f_2}{\partial \lambda_1} &  \ddots &  &\vdots  \\
\vdots  &  & \ddots  &\vdots \\
\frac{\partial f_m}{\partial \lambda_1} &  \frac{\partial f_m}{\partial \lambda_2} & \cdots & \frac{\partial f_m}{\partial \lambda_d} 
\end{array}
\right]
\end{align*}
\end{definition}
%The above definition matches the spirit discussed in \citet{Magnus10} on generalization of vector derivative to matrix derivative. 
Now, if the vector differential operator is defined as a column vector
%Using vector differential operator 
  $\mathbf{D}_{\bs{\lambda}} = \left( \frac{\partial}{\partial \lambda_1}, \frac{\partial}{\partial \lambda_2},\ldots, \frac{\partial}{\partial \lambda_d} \right)'$, then 
  the Jacobian matrix can be re-written as: 
$$\mathbf{J}(\mathbf{f}(\bs{\lambda})) = \mathbf{D}_{\bs{\lambda}}(\mathbf{f}) = \mathbf{f}(\bs{\lambda})\mathbf{D}'_{\bs{\lambda}}   = \left( f_1(\lambda), f_2(\lambda), \ldots, f_m(\lambda) \right)' \left( \frac{\partial}{\partial \lambda_1}, \frac{\partial}{\partial \lambda_2},\ldots, \frac{\partial}{\partial \lambda_d} \right)  $$
This implies that to match the definition of derivative from matrix calculus, the vector differential operator should be applied from right to the left. %This is also a requirement to be satisfied on new definition , as discussed by \citet{Magnus10} on generalization of vector derivative to matrix derivative. 
This is also to fulfill the requirement  that a new defin
So, applying vector derivative operator from right to the left,  has been kept as a rule throughout the article. 
%Let us define the derivative operator as a vector $\mathbf{D}_\mathbf{x} = (\frac{\partial}{\partial x_1}, \frac{\partial}{\partial x_2},\ldots, \frac{\partial}{\partial x_d})'$.

%\hspace{0.2 in} The specific K-derivative operator for Jacobian in vector form is obtained as under:%The followng  rules for differentiation of a Kronecker product are useful. 
\begin{definition}[The K-derivative Operator] 
Let $\bs{\lambda} = (\lambda_1, \lambda_2, \ldots, \lambda_d)'$, $\bs{\lambda} \in \mathbb{R}^d$, the vector differential operator $\mathbf{D}_{\bs{\lambda}} = \left( \frac{\partial}{\partial \lambda_1}, \frac{\partial}{\partial \lambda_2},\ldots, \frac{\partial}{\partial \lambda_d} \right)'$ and  $\mathbf{f}(\bs{\lambda}) = (f_1(\lambda), f_2(\lambda), \ldots, f_m(\lambda))' \in \mathbb{R}^{m}$ be a differentiable m-component vector function.  Then, the  Kronecker product between $\mathbf{D}_{\bs{\lambda}}$ and $\mathbf{f}(\bs{\lambda})$ is given as under:
%K-derivative operator $\mathbf{D}_{\bs{\lambda}}^\otimes $ is defined as:
\begin{align*}
\mathbf{D}_{\bs{\lambda}}^\otimes \mathbf{f}(\bs{\lambda}) &= 
%Vec\left(\frac{\partial f}{\partial \bs{\lambda}'}\right)' = Vec(\frac{\partial}{\partial \bs{\lambda}}f') = 
Vec\left[
\begin{array}[]{c c c c}
\frac{\partial f_1}{\partial \lambda_1} & \frac{\partial f_1}{\partial \lambda_2}  & \cdots & \frac{\partial f_1}{\partial \lambda_d} \\
\frac{\partial f_2}{\partial \lambda_1} &  \ddots &  &\vdots  \\
\vdots  &  & \ddots  &\vdots \\
\frac{\partial f_m}{\partial \lambda_1} &  \frac{\partial f_m}{\partial \lambda_2} & \cdots & \frac{\partial f_m}{\partial \lambda_d} 
\end{array}
\right]' \\
\Rightarrow \mathbf{D}_{\bs{\lambda}}^\otimes \mathbf{f}(\bs{\lambda}) &= Vec\left(\frac{\partial \mathbf{f}}{\partial \bs{\lambda}'}\right)' = Vec \left( \frac{\partial}{\partial \bs{\lambda}}\mathbf{f}' \right)
\end{align*}
where, the $Vec$ operator converts $m \times d$ matrix into an $ md \times 1$ column vector by stacking the columns one after an other. 
The operator $\mathbf{D}_{\bs{\lambda}}^{\otimes } $ is called Kronecker derivative operator  or simply, K-derivative operator. % $\mathbf{D}_{\bs{\lambda}}^\otimes $ is defined as:
\end{definition} 
%The term $\frac{\partial}{\partial \bs{\lambda}'}\mathbf{f}$ is the Jacobian (first order derivative) of $\mathbf{f}(\bs{\lambda})$. 
Thus, the Kronecker product with the vector differential operator, obtains vectorization of the transposed Jacobian of a vector function.  Corresponding to the definition, the k-th order differentiation is given by:
%\begin{align*}\end{align*}
%\begin{align*} 
$$ \mathbf{D}_{\bs{\lambda}}^{\otimes k} \mathbf{f} = \mathbf{D}_{\bs{\lambda}}^{\otimes}\left( \mathbf{D}_{\bs{\lambda}}^{\otimes k-1}\mathbf{f} \right) = [f_1(\lambda), f_2(\lambda), \ldots, f_m(\lambda)]' \otimes \left[ \frac{\partial }{\partial \lambda_1},  \frac{\partial }{\partial \lambda_2},  \cdots,  \frac{\partial }{\partial \lambda_d}  \right]^{'\otimes k}  $$
%\end{align*}
The $\mathbf{D}_{\bs{\lambda}}^{\otimes k}\mathbf{f}$ is a column vector of dimension $md^k \times 1$. 

\hspace{0.2 in} Some important properties of the K-derivative operator are listed below. 
%For that let us define the derivative operator as a vector $\mathbf{D}_\mathbf{x} = (\frac{\partial}{\partial x_1}, \frac{\partial}{\partial x_2},\ldots, \frac{\partial}{\partial x_d})'$.
\begin{definition}[Scaling Property] 
Let $\bs{\lambda} = (\lambda_1, \lambda_2, \ldots, \lambda_d)'$, $\bs{\lambda} \in \mathbb{R}^d$, $\mathbf{f}(\bs{\lambda}) \in \mathbb{R}^{m}$ and $\mathbf{f}_1(\bs{\lambda}) = \mathbf{A}\mathbf{f}(\bs{\lambda})$, where $\mathbf{A}$ is an $n \times m$ matrix. Then 
\begin{align*}
\mathbf{D}_{\bs{\lambda}}^\otimes (\mathbf{f}_1) = \left( \mathbf{A} \otimes \mathbf{I}_d \right) \mathbf{D}_{\bs{\lambda}}^\otimes (\mathbf{f})
\end{align*}
where, $\mathbf{I}_d$ is a d-dimensional unit matrix.
\end{definition} 

\begin{definition}[Chain Rule] 
Let $\bs{\lambda} = (\lambda_1, \lambda_2, \ldots, \lambda_d)'$, $\bs{\lambda} \in \mathbb{R}^d$, $\mathbf{f}(\bs{\lambda}) \in \mathbb{R}^{m_1}$  and $\mathbf{g}(\bs{\lambda}) \in \mathbb{R}^{m_2}$. Then 
\begin{align*}
\mathbf{D}_{\bs{\lambda}}^\otimes (\mathbf{f} \otimes \mathbf{g}) = \mathbf{K}^{-1}_{3 \leftrightarrow 2}(m_1, m_2, d)((\mathbf{D}_{\bs{\lambda}}^\otimes \mathbf{f})\otimes \mathbf{g}) + \mathbf{f} \otimes (\mathbf{D}_{\bs{\lambda}}^\otimes \mathbf{g})
\end{align*}
where, %$\mathbf{D}_{\bs{\lambda}}$ is the derivative operator as a vector $\mathbf{D}_{\bs{\lambda}} = (\frac{\partial}{\partial \lambda_1}, \frac{\partial}{\partial \lambda_2},\ldots, \frac{\partial}{\partial \lambda_d})'$  and
$\mathbf{K}_{3 \leftrightarrow 2}(m_1, m_2, d)$ is a commutation matrix of size $m_1m_2d \times m_1m_2d$ that changes the order of the the Kronecker product components. For example, 
\begin{align*}
\mathbf{K}_{3 \leftrightarrow 2}(m_1, m_2, d)(\mathbf{a_1}\otimes \mathbf{a_2} \otimes \mathbf{a_3} ) =  \mathbf{a_1} \otimes \mathbf{a_3} \otimes \mathbf{a_2}
\end{align*}
\end{definition}
%Another important derivative rule is:
The application of Scaling property gives a following simplification: 
%\begin{definition}[Rule 2]
%Let $\bs{\lambda} = (\lambda_1, \lambda_2, \ldots, \lambda_d)'$, $\bs{\lambda} \in \mathbb{R}^d$, and $\mathbf{x}$ is a vector of constants. Then:
\begin{align*}
\mathbf{D}_{\bs{\lambda}}^{\otimes }{\bs{\lambda}}^{\otimes k}  = \left( \sum_{j=0}^{k-1} \mathbf{K}_{j+1 \leftrightarrow k}( d_{[k]} ) \right)\left( {\bs{\lambda}}^{\otimes (k-1)}\otimes \mathbf{I}_d\right),  
\end{align*}
where, $d_{[k]}= [d,d,\ldots,d]'_{(k \times 1)}$.
%\end{definition}
%Also,
The repeated applications of chain rule gives a following simplification: 
%\begin{definition}[Rule 3] 
\begin{align*}
\mathbf{D}_{\bs{\lambda}}^{\otimes r}\mathbf{a}'^{\otimes k}{\bs{\lambda}}^{\otimes k} = k(k-1) \cdots (k-r+1)\left[(\mathbf{a'}{\bs{\lambda}})^{k-r}\mathbf{x}^{\otimes r} \right]
\end{align*}
where, $\mathbf{a}$ is a vector of constants. 
\vskip 0.2in
\bibliography{Allref}

\end{document}